\documentclass{article}

\usepackage{arxiv}

\usepackage[utf8]{inputenc} 
\usepackage[T1]{fontenc}    
\usepackage{hyperref}       
\usepackage{url}            
\usepackage{booktabs}       
\usepackage{amsfonts}       
\usepackage{nicefrac}       
\usepackage{microtype}      
\usepackage{lipsum}		
\usepackage{graphicx}
\usepackage[square,numbers]{natbib}
\usepackage{doi}

\usepackage{longtable}
\usepackage{multirow}

\title{Rule-Extraction Methods From Feedforward Neural Networks: A Systematic Literature Review}

\author{Sara ~El Mekkaoui\\
	Équipe AMIPS\\
	École Mohammadia d’Ingénieurs\\
	Rabat, Morocco\\
	\texttt{saraelmekkaoui@research.emi.ac.ma}\\
	\And
	Loubna ~Benabbou \\
	Département Sciences de la Gestion\\
	Université du Québec à Rimouski, Lévis\\
	Qc, Canada\\
	\texttt{loubna\_benabbou@uqar.ca} \\
        \AND
	Abdelaziz ~Berrado \\
	Équipe AMIPS\\
	École Mohammadia d’Ingénieurs\\
	Rabat, Morocco \\
	\texttt{berrado@emi.ac.ma} \\
}

\renewcommand{\shorttitle}{Under Review}

\hypersetup{
pdftitle={Rule-Extraction Methods From Feedforward Neural Networks: A Systematic Literature Review},
pdfsubject={},
pdfauthor={Sara~El Mekkaoui, Loubna~Benabbou, Abdelaziz~Berrado},
pdfkeywords={},
}

\begin{document}
\maketitle

\begin{abstract}
Motivated by the interpretability question in ML models as a crucial element for the successful deployment of AI systems, this paper focuses on rule extraction as a means for neural networks interpretability. Through a systematic literature review, different approaches for extracting rules from feedforward neural networks, an important block in deep learning models, are identified and explored. The findings reveal a range of methods developed for over two decades, mostly suitable for shallow neural networks, with recent developments to meet deep learning models' challenges. Rules offer a transparent and intuitive means of explaining neural networks, making this study a comprehensive introduction for researchers interested in the field. While the study specifically addresses feedforward networks with supervised learning and crisp rules, future work can extend to other network types, machine learning methods, and fuzzy rule extraction.
\end{abstract}


\section{Introduction}\label{sec1}

The growing deployment of Artificial Intelligence (AI) systems in sensitive domains and the emerging regulatory and ethical considerations surrounding AI have fostered the development of Machine Learning (ML) models' interpretability. This area of research has gained significant attention in recent years \cite{vilone2020}, especially with the emergence of complex models in real-world applications. 

Feedforward neural networks play an essential role in Deep Learning (DL), serving as the fundamental building blocks for numerous advanced models. These networks are important because they can process and transform input data through multiple layers of interconnected neurons, enabling the extraction of useful representations. Their hierarchical structure allows for the construction of deep neural networks, which have demonstrated remarkable success in various domains, including computer vision, natural language processing, and speech recognition. While DL models have succeeded in numerous fields, the drive for improved accuracy has led to the development of increasingly complex architectures and techniques. The opacity of their internal mechanisms poses significant challenges in understanding their decision-making process, particularly in critical domains such as healthcare, finance, and safety-critical applications. Addressing the interpretability challenge in DL is a pressing concern, as it not only improves the understanding of the model but also enables fairness and ethical evaluation, enhancing control and robustness and discovering insightful knowledge.

While interpretability in ML, especially in the context of neural networks, is often considered a relatively recent area of research, its origins can be traced back to the late 1980s with the advent of neural-symbolic computing \cite{garcez2019}. Neural symbolic systems represent an interdisciplinary approach combining principles from neural networks and symbolic artificial intelligence. These systems aim to integrate the powerful learning capabilities of neural networks with the interpretability and reasoning abilities offered by symbolic reasoning. A fundamental aspect of neural-symbolic systems lies in extracting knowledge from neural networks in the form of rules, leading to the development of various rule extraction techniques from neural networks. Notably, these approaches have similarities with contemporary methods employed for DL explainability \cite{townsend2019}. The significance of these techniques has even been discussed in major conferences, such as the NIPS and ICML, with a dedicated NIPS 96 Workshop, reflecting their relevance within the scientific community.

Motivated by the significant contributions in rule extraction from neural networks, this paper contributes a systematic review of rule extraction methods from feedforward neural networks employed in supervised learning tasks. The study offers a comprehensive and extended taxonomy incorporating contemporary classification criteria for explainability and a novel classification according to different approaches. Furthermore, it addresses the transferability of existing rule extraction techniques to deep feedforward networks. The comprehensive nature of this survey renders it invaluable to researchers interested in this field, as it consolidates many studies within a single research.

The contribution of this study can be summarised as follows:

\begin{itemize}
 \item A systematic review of papers on rule extraction methods from feedforward neural networks.
 \item An extended taxonomy including some modern explainability techniques classification criteria and a categorization of the rule extraction approaches offering a comprehensive overview of available techniques.
 \item A \href{https://saraelmekkaoui.github.io/rule_ext_ann/}{visual literature browser} to improve information accessibility using SurVis \cite{beck2016} 
\end{itemize}

This paper is organized as follows: Section \ref{sec2} presents the main concepts and definitions. Section \ref{sec3} provides details of the review methodology and descriptive findings. Section \ref{sec4} gives the main classification framework and details the dimension related to the rule extraction approaches categories. Section \ref{sec5} discusses the challenges of using these methods for DL models, and Section \ref{sec6} concludes this paper.

\section{Background and main concepts}\label{sec2}

\subsection{Feedforward Neural Networks}\label{sec2-1}

\begin{figure}[ht]
\centering
\includegraphics[width=0.8\textwidth]{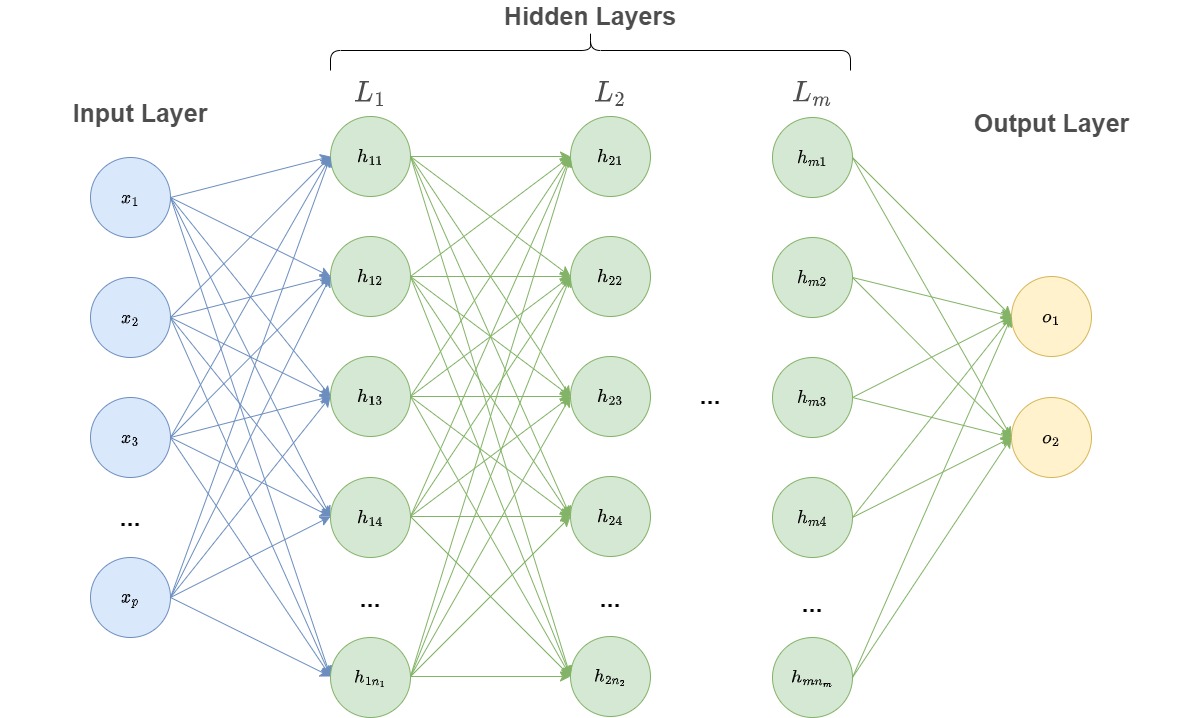}
\caption{A representation of a deep feedforward neural network}\label{neural_network_representation}
\end{figure}

DL models are nonlinear parametric ML models inspired by neuroscience \cite{goodfellow2016}. Previously known as connectionism or artificial neural networks, the modern term DL represents today a general principle of learning multiple levels of composition. Deep feedforward networks are typical DL models called Multilayer Perceptron (MLP). In a supervised learning setting, the objective of a deep feedforward network is to map an input $x$ to an output $o$. The output can be a real number in the case of a regression problem or a category in the case of a classification problem. Deep feedforward networks, hereafter referred to as networks, are composed of processing elements called neurons, nodes, or units organized as a network as depicted in Figure \ref{neural_network_representation}. The units are spread between multiple layers. The input layer $L_0$ receives the input features, followed by the first hidden layer $L_1$. Each hidden layer, including the first, consists of a specific number of units. Each unit $u_{ij}$ from layer $L_j$ receives inputs from the previous layer $L_{j-1}$, performs a linear combination of these inputs then computes an activation value. The last layer is the output layer that provides predictions. The most popular activation functions are sigmoid, rectifier linear unit, and hyperbolic tangent functions \cite{rasamoelina2020}. 

In a supervised learning setting, fitting a network refers to learning the parameter values, i.e., weights and biases, that result in a low error between the network predictions and the true outputs. Backpropagation gradient descent is the most used learning algorithm consisting of propagating the error backward through the network and updating its parameters weights to reduce the gradient [ Ref]. The error is measured using a loss function. The most popular loss functions are the cross entropy loss in classification problems and the mean squared error in regression problems. Effective development of DL models is a highly empirical process that requires searching for the best hyper-parameters such as the number of hidden layers, number of hidden units, learning rate, number of iterations, and activation function. Different architectures could be validated using a validation dataset to determine the best configuration \cite{raschka2018}.

\subsection{Rule extraction from feedforward neural networks}\label{sec2-2}

\subsubsection{Rule extraction problem}\label{sec2-2-1}
The problem of rule extraction from feedforward neural networks consists of generating comprehensible hypotheses in the form of symbolic logical representations from a trained network. For instance, in the case of Boolean inputs $x_i$ used for a classification task with two classes $o_1$ and $o_2$, a rule extraction method providing propositional rules would generate rules in the format of:

IF ($x_1 \wedge \neg x_2 \wedge x_5$) THEN $o_1$

Most existing rule extraction methods focus on networks with discrete inputs and outputs. Inputs are considered Boolean or nominal valued, and outputs as classes. In cases where inputs or outputs are continuous, a discretization step is typically required.

Various approaches have been proposed for rule extraction from feedforward neural networks, and previous literature reviews in this field have provided classification schemes to categorize these methods. The diversity of these approaches illustrates the significant research effort invested in this area.

\subsubsection{Overview of previous surveys}\label{sec2-2-2}
Rule extraction from neural networks has been an area of research since the late 80s. Table \ref{tab:rule_ext_def} defines the main concepts discussed in this section, and Table \ref{tab:surveys_list} summarises the existing surveys on rule extraction methods.  The ADT taxonomy, referring to the authors' initials, for rule extraction methods was proposed in 1995 \cite{andrews1995} and refined in 1998 \cite{tickle1998}. It follows five dimensions:
\begin{itemize}  
\item  The expressive power of the extracted rules: we refer to this dimension as \textit{rules form} to indicate the type of the rules extracted from the network, such as Boolean, fuzzy, and first-order logic rules.
\item Translucency into the network: this dimension includes three classes of extraction methods as pedagogical, decompositional, and eclectic. They characterize the infiltration of the rule extraction method into the internals of the network. This dimension is similar to the modern model-related dimension of agnostic or specific methods.
\item Portability of the technique across neural network architectures and training regimes: this dimension is about the requirements of the extraction methods for a particular training regime to simplify the extraction task. It is similar to the modern interpretability techniques characteristic of being intrinsic or post-hoc.
\item The quality of the extracted rules: including accuracy, fidelity, consistency, and comprehensibility metrics to assess the performance of the extracted rules.
\item The complexity of the rule extraction algorithm: characterizing the efficiency of the rule extraction process. 
\end{itemize}

\begin{longtable}{p{2.7cm}p{2.7cm}p{10cm}}
\caption{Definitions of related concepts to rule extraction methods from feedforward neural networks}\label{tab:rule_ext_def}\\
\hline
Dimension & Concept & Definitions \\
\hline
\multirow[t]{3}{2cm}{Rules form} 
& Propositional rules & The conditional part contains conjunctions, disjunctions, and negations. It is expressed as a combination of Boolean conditions. Examples of propositional rules in the case of discrete features and a classification task could be "if $X=1$ and $Y=0$ then class 1". For continuous variables, values are restricted in forms like "if $X \in [a,b], X \leq a$ with $a, b \in \mathbb{R}$". \\
\cline{2-3}
& MofN rules & They are Boolean expressions specified by the integer threshold $m$ and a set of $n$ Boolean conditions. They can take the form "if at least/exactly/at most $m$ of the $n$ conditions are satisfied then class 1". They can be converted into propositional rules.\\ 
\cline{2-3}
& Obliques rules & They express nonparallel decision boundaries with input axes as linear inequalities "if $aX_1+bX_2 < c$ and $dX_4+eX_3 > f$  then class 0, with $a, b, c, d, e, f \in \mathbb{R}$".\\
\cline{2-3}
& Fuzzy rules & They are expressed in linguistic concepts, e.g., "if X is high and Y is low then class 1".\\  
\hline
\multirow[t]{4}{2cm}{Translucency into the network} 
& Pedagogical & The methods target the function learned by the network and treat the network as a black box. The extracted rules describe the behavior of the network output in terms of inputs without using the network internals.\\
\cline{2-3}
& Decompositional & The methods target the network internals by extracting rules at each hidden and output unit and then aggregating them into global rules linking inputs to outputs.\\
\cline{2-3}
& Eclectic & The methods are hybrid, examining the network units but extracting rules describing the relationship between the inputs and the outputs directly. \\
\hline

\multirow[t]{2}{2cm}{Portability}
& Constrained & The extraction methods require a modification of the network or a  special training regime.\\ 
\cline{2-3}
& Unconstrained & The extraction methods do not impose any changes to the network structure or training regime.\\
\hline
\multirow[t]{5}{2cm}{Rules quality} 
& Accuracy / Correctness & Measuring the extracted rules predictions accuracy with regards to the true targets.\\
\cline{2-3}
& Fidelity & Measuring the prediction accuracy with regards to the network prediction. This measures how well the extracted rules are in mimicking the network behavior.\\
\cline{2-3}
& Consistency & Measuring how consistent the extracted rules predictions are under different network training sessions.\\
\cline{2-3}
& Comprehensibility & Measuring the ability to understand the extracted rules using their size, i.e., number of rules and antecedents per rule or average rule length. \\
\cline{2-3}
& Completeness & Characterizing the extracted rules power in terms of the ratio of input examples covered by the extracted rule over total input examples.\\
\hline
Complexity & \multicolumn{2}{p{9cm}}{The complexity of the extraction method, which for instance, could be exponential or polynomial.} \\
\hline
\multirow[t]{2}{2cm}{Application}
& Model-agnostic & The extraction method can be applied to any ML technique.\\ 
\cline{2-3}
& Model-specific & The extraction method is specific to feedforward neural networks. \\
\hline
\multirow[t]{2}{2cm}{Design}
& Intrinsic & The extraction method requires a simplification of the network complexity using techniques such as regularization.\\ 
\cline{2-3}
& Post-hoc & The extraction method intervenes after the network training. \\
\hline
\multirow[t]{2}{2cm}{Scope}
& Global & The extraction method offers an explanation of the entire network behavior.\\ 
\cline{2-3}
& Local & The extraction methods give an explanation of the decision related to a particular input example.  \\
\hline
\end{longtable}

The proposed taxonomy has various dimensions indicating the diversity of the proposed extraction methods during the 90s. Table \ref{tab:surveys_list} shows the existing surveys on rule extraction methods from neural networks. Surveys following \cite{andrews1995,tickle1998} mainly classify papers using the translucency dimension as pedagogical, decompositional, or eclectic. However, the existing classification scheme does not offer a clear overview of the nature of methods used to extract rules from networks. Unlike recent reviews, this study performs an extensive review and classifies the existing literature following a comprehensive dimension describing the main approaches to rule extraction. The following section presents the review methodology and main descriptive findings.

\begin{table}[h]%
\centering %
\caption{Surveys of rule extraction methods from feedforward neural networks}\label{tab:surveys_list}
\begin{tabular}{p{1.5cm}p{6.5cm}p{7cm}}
\hline
Reference & Title & Classification scheme\\
\hline
\cite{andrews1995} & Survey and critique of techniques for extracting rules from trained artificial neural networks   & ADT classification of rule extraction methods. \\
\hline
\cite{craven1996} & Extracting comprehensible models from trained neural networks & Classification into Global (pedagogical) and local (decompositional and eclectic).\\
\hline
\cite{tickle1998} & The truth will come to light: Directions and challenges in extracting the knowledge embedded within trained artificial neural networks & Extension of the ADT classification. \\
\hline
\cite{huysmans2006} & Using Rule Extraction to Improve the Comprehensibility of Predictive Models & Classification into dependent and independent methods from the network. \\
\hline
\cite{augasta2012} & Rule extraction from neural networks — A comparative study & Classification into pedagogical, decompositional, and eclectic.  \\
\hline
\cite{hailesilassie2016} & Rule Extraction Algorithm for Deep Neural Networks: A Review & Classification into pedagogical, decompositional, and eclectic. \\
\hline
\cite{he2020} & Extract interpretability-accuracy balanced rules from artificial neural networks: A review & Classification into pedagogical, decompositional, and eclectic. \\
\hline
\end{tabular}
\end{table}

\section{Literature Review}\label{sec3}

\subsection{Review methodology}\label{sec3-1}
This study conducted a systematic literature review based on the guidelines suggested by \cite{kitchenham2004}. The methodology includes three steps of planning, conducting, and reporting the review. The first step is planning the review by identifying the research area and developing a research protocol to guide the review process. The review protocol involves the following steps: definition of the research area, research questions, search procedure, inclusion and exclusion criteria, quality assessment, data extraction, and data synthesis. 

This review's research scope is "rule extraction methods from feedforward neural networks for supervised learning", with a focus on crisp rules. The review covers techniques for extracting rules from both shallow and deep feedforward neural networks. This study has four goals: (i) identify and introduce a comprehensive and exhaustive classification framework, (ii) identify the trends, (iii) describe the different approaches to rule extraction methods, and (iv) identify which approaches are suitable for DL models. To achieve the objective of the review, the following research questions were considered:

\begin{itemize}
    \item \textbf{RQ1} What classification dimensions could be considered for rule extraction methods?
    \item \textbf{RQ2} What are the research trends in this field?
    \item \textbf{RQ3} What are the main approaches to rule extraction from feedforward neural networks?
    \item \textbf{RQ4} What approaches are applicable in the case of DL models?
    
\end{itemize}

The inclusion/exclusion criteria are applied to ensure that the selected studies are within the review scope. The main selection criteria are (i) presenting a method for rule extraction from feedforward networks and (ii) being written in the English language. The exclusion criteria include (i) applying rule extraction methods and (ii) studies on rule extraction methods from other ML techniques. The search process was conducted by examining papers from major academic databases and the previous surveys listed in Table \ref{tab:surveys_list}. Search terms related to “rule extraction” and “neural networks” and their common variations were used to select papers. As we are interested in the whole evolution of this field, no restrictions were imposed on the publication year. The extracted papers were scanned based on their titles, abstracts, and keywords to determine if they met the selection criteria. Some papers also required scanning the introduction and/or the conclusion sections. The obtained list was completed with papers from the surveys. The selected papers' “review” or “related work” sections were scanned to complete the list of relevant papers. 

A detailed scan of the obtained papers was conducted in order to assess their quality. Papers with clear methodology descriptions and experiments to showcase the results of the proposed method were chosen. In order to identify classification dimensions and main approaches, a detailed analysis of each study was conducted, and a synthesis describing the proposed method was generated. An initial extraction form was developed to capture relevant information, including publication year, authors, publication venue, translucency type (pedagogical, decompositional, or eclectic), and method portability. The extraction form was refined during the data synthesis phase to ensure comprehensive data collection and analysis. The final list includes 89 papers. 

\subsection{Taxonomy}\label{sec3-2}

\begin{figure}[ht]
\centering
\includegraphics[width=1\textwidth]{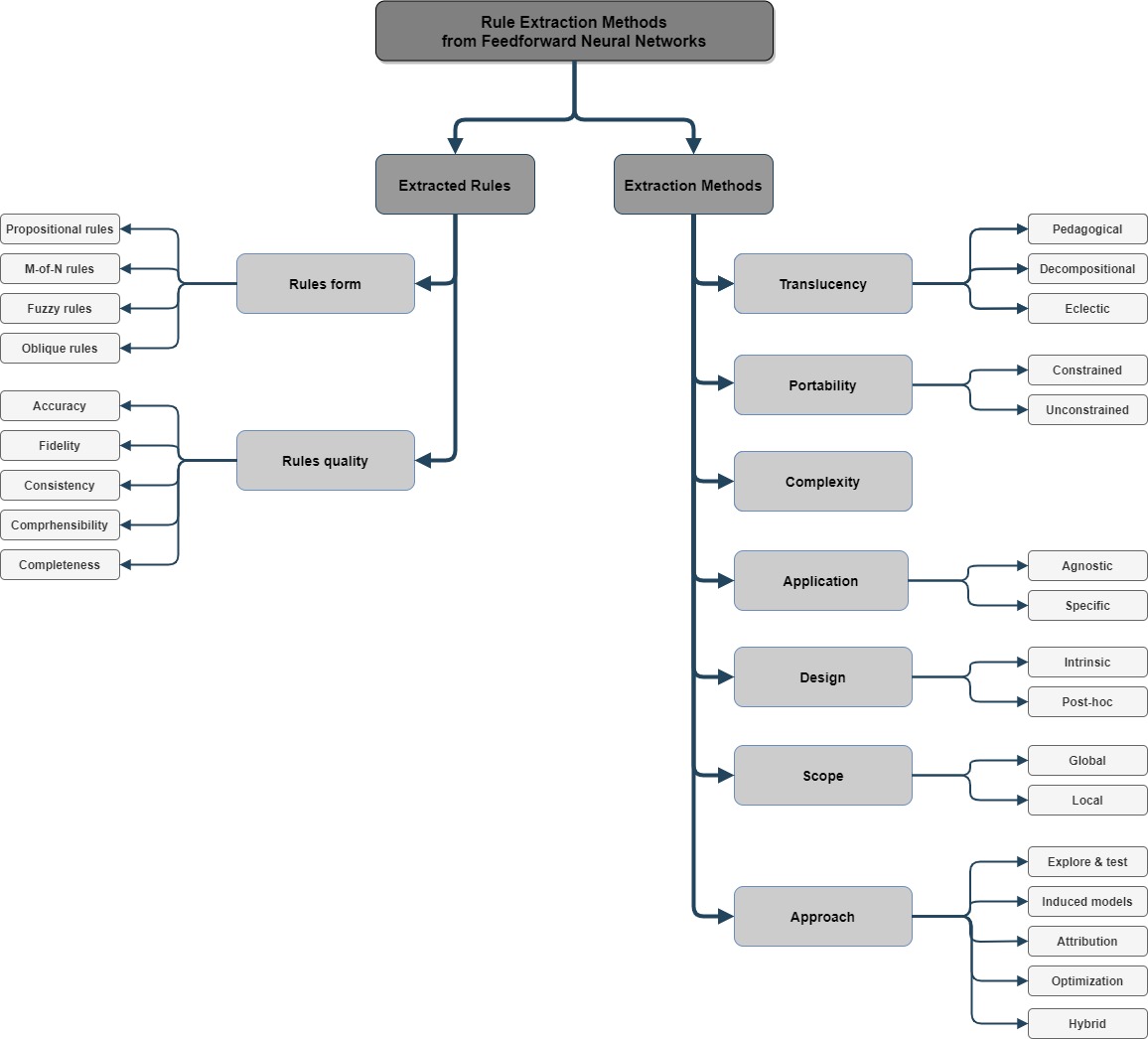}
\caption{Taxonomy of rule extraction methods from feedforward neural networks}\label{fig:rule_ext_classification_framework}
\end{figure}

Figure \ref{fig:rule_ext_classification_framework} illustrates the taxonomy for rule-extraction methods from feedforward neural networks. The taxonomy dimensions are defined in Table \ref{tab:rule_ext_def}.

The classification dimensions can be related to the extraction method or the extracted rules. The translucency dimension is specific to rule extraction methods from feedforward neural networks, describing how the extraction is performed. Pedagogical methods extract rules that describe the behavior of output units in terms of input units. Decompositional methods extract rules at each hidden and output unit, with the concept associated with each unit being the rule consequent and certain subsets of units feeding into this unit representing the antecedents of the rule. The process of rule extraction involves finding sufficient conditions for each consequent. Decompositional rule extraction methods assume that hidden and output unit activation values are discrete after training. The extracted rules are then aggregated to form rules that describe the relationship between inputs and outputs. Eclectic methods lie between the pedagogical and decompositional types, as they can analyze the internals of the network but extract global rules between inputs and outputs.

The portability dimension, proposed by Andrews et al. \cite{andrews1995}, characterizes how dependent an extraction method is on the network. Portable methods do not make any assumptions about the network and do not use any internal information, and can be used with any network. The portability dimension can be seen as a special case of the modern interpretability dimension of application, separating interpretability tools into model-agnostic or model-specific tools.

The application, complexity, design, and scope dimensions can be seen as general classification criteria for any interpretability tool. The complexity of the rule extraction method can vary depending on the algorithm used, the number of rules being extracted, and their complexity. The complexity of the rule extraction methods may also depend on the complexity of the network architecture or the high dimensionality of the data being processed. The design of the extraction method can be intrinsic or post-hoc. Intrinsic methods aim to simplify the network complexity to facilitate rule extraction, and regularization and pruning techniques can be used to make the network more interpretable and help in efficiently extracting rules. Post-hoc techniques are applied after the network is completely trained and do not impose any training requirements. The scope can be global when the extracted rules offer a general explanation of the network or local when the explanations are specific to certain samples.

This review introduces a dimension related to the employed approach to categorize rule extraction methods from a technical perspective. In Section \ref{sec4}, we will describe in detail the various approaches to building rules, including explore \& test, induced models, attribution, optimization, and hybrid  methods. The extracted rules can take different forms, and their quality can be assessed using different metrics, as described in detail in Table \ref{tab:rule_ext_def}.

\subsection{Descriptive findings}\label{sec3-3}

\begin{figure}[ht]
\centering
\includegraphics[width=0.9\textwidth]{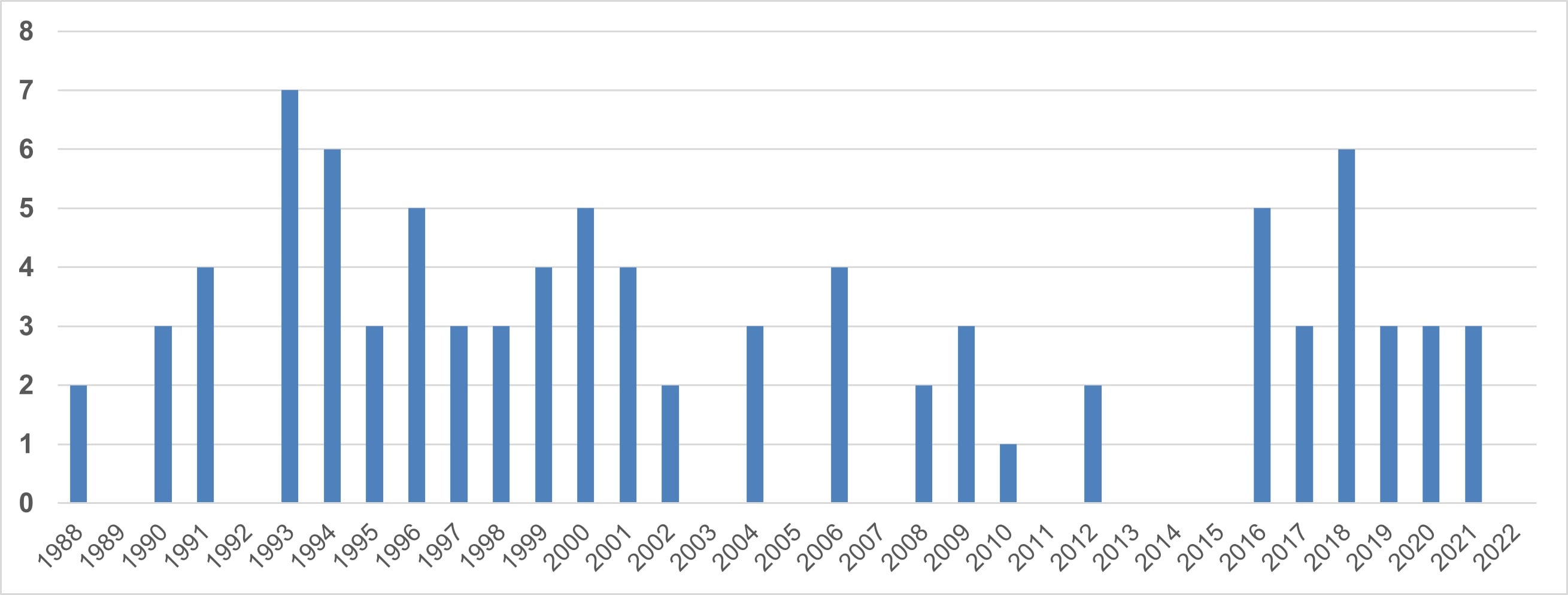}
\caption{Evolution of studies on rule extraction methods from feedforward neural networks expressed by the number of studies per year}\label{fig:studies_chrono}
\end{figure}

Figure \ref{fig:studies_chrono} illustrates the distribution of the selected studies over time. The first rule extraction method was introduced in 1988, followed by a significant number of papers during the 1990s. However, the research field declined thereafter, possibly due to reduced interest in neural networks. However, there has been renewed interest in the field in recent years, as the interpretability of neural networks has resurfaced recently as an important area of research.

Out of the 89 papers, some have proposed multiple methods within the same paper, while others have proposed the same method in different papers. There are 81 distinct methods, of which 78 are global and 3 are local. 

\begin{figure}[ht]
\centering
\includegraphics[width=0.7\textwidth]{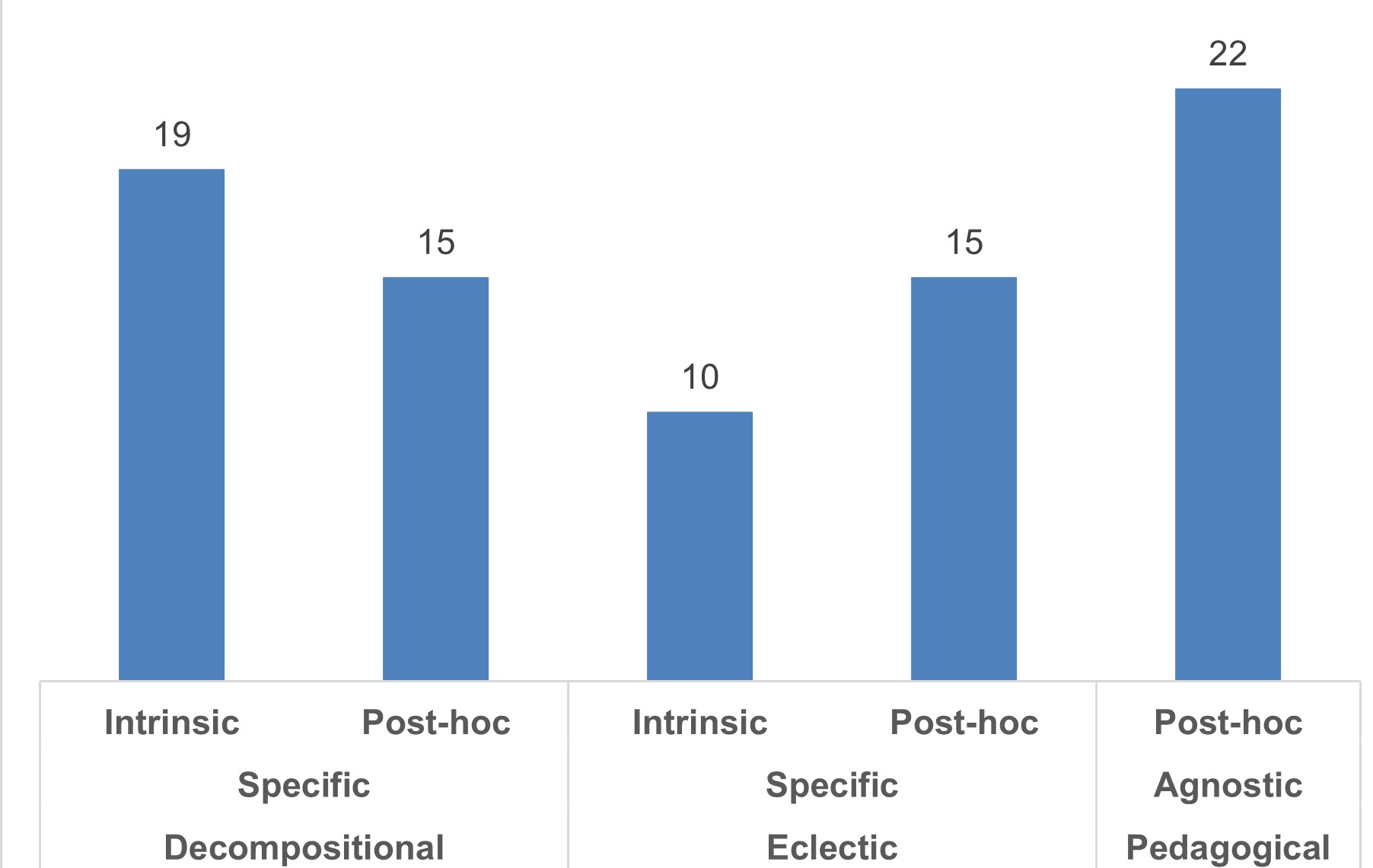}
\caption{Distribution of the rule extraction methods from feedforward neural networks following the dimensions "Translucency", "Application", and "Design"}\label{fig:descriptive_results}
\end{figure}

Figure \ref{fig:descriptive_results} depicts the distribution of 81 methods based on the dimensions of "Translucency", "Application", and "Design". It is worth noting that Pedagogical approaches are inherently agnostic and post-hoc, as they do not rely on any assumptions about the network or training regime and do not utilize any information about the network internals. However, when a method requires access to the internal workings of the network, it may fall into the categories of eclectic or decompositional, and in both cases, these methods are specifically designed for neural networks. Eclectic and decompositional methods can be considered intrinsic if they directly affect the network internals or post-hoc if they operate after the training.

\section{Rule extraction approaches}\label{sec4}

The great majority of rule extraction techniques for feedforward neural networks can be regarded as proxy methods, which are interpretability strategies employed to clarify the behaviour of black-box models. A surrogate transparent model is developed through these approaches by training a simpler model, such as linear regression or decision trees, using the same input features as the black-box model and its predictions as targets. In the context of rule extraction, the mentioned proxy approach is pedagogical. However, it can also be decompositional if the proxy is fitted at the hidden and output units level. The surrogate model acts as a substitute or approximation of the unit activations, facilitating comprehension of its predictions. In the context of proxies, rules may be extracted by performing a search procedure through the space of rules and testing them against the network outputs or by utilizing induced models like decision trees. Additionally, we discuss other methods of extracting rules, including attribution methods based on feature importance and optimization methods involving defining the input space that maximizes a specific output class. Finally, some of the proposed methods combine different techniques and are considered as hybrid.


\subsection{Explore \& test}\label{sec4-1}

Early work on neural-symbolic computing has motivated the extraction of rules from neural networks using the explore \& test approach. It follows a typical rule learning process by exploring the space of possible rules and testing them against the network outputs. Table \ref{tab:explore_test} summarises the explore \& test methods. They all provide global explanations. 

Typically, learning rules can be formulated as a search problem \cite{furnkranz2012} defined by (i) an appropriate search space, (ii) a search strategy for searching for rules through this space, and (iii) a quality function to evaluate the rules: 
\begin{itemize}
    \item Search space: enumerating the whole space of possible rules is often infeasible. As exploring the rules’ space might be intractable, organizing and structuring the search space is necessary. Hence, rules extraction from neural networks is constructed by searching for the conjunction of features that is most predictive for a network response. Exhaustive, heuristic, or stochastic algorithms perform the search.
    \item Search strategy: can be organized following the general-to-specific, specific–to –general, or bi-directional approaches. The general-to-specific rule search strategy is a top-down approach starting from the most general rule and further specializing it. The specific-to-general rule search or the growing strategy is a bottom-up approach that looks for specific rules, for instance, an input example, before finding more general rules. The bi-directional approach combines both of the previous approaches.
    \item Rules evaluation: the quality of rules can be evaluated using different measures such as precision and information gain. The basic principle underlying many quality measures are consistency and coverage. 
\end{itemize}

A widely adopted approach in rule extraction entails the utilization of subset-type algorithms \cite{towell1993} to extract rules in a decompositional manner. This methodology aims to derive rules at both the hidden and output unit levels, employing discrete activation values to simplify the extraction process. Subset-type algorithms typically involve three main steps:

\begin{itemize}
    \item Exploration of unit input combinations: The algorithm searches for different combinations of inputs whose weighted sum activates the unit for each network unit.
    \item Rule generation: Upon identifying each combination of inputs that activates a unit, the algorithm generates a rule wherein the input to the specific combination of links defines the premises. All premises of a rule are conjuncted. 
    \item Rule merging: The extracted rules are subsequently merged to establish direct linkages between inputs and outputs within the framework of decompositional methods. This merging process yields rules that directly associate input conditions with the corresponding output decisions.
\end{itemize}

A breadth-first subset algorithm is a simple approach to searching for subsets of incoming weights that activate a unit. It starts with sets containing one link, verifies if they activate the unit, and rewrites valid sets as rules. Then, it increases the size of the subset until all subsets have been explored. Finally, it removes redundant rules. 

However, the major problem with these methods lies in the search complexity. For instance, if a network unit has $n$ inputs with binary values, there could be as many as $2^n$ distinct input sets. Another challenge is how to choose an appropriate manner for activation values discretization. Therefore, some methods make assumptions to narrow the search of the rules space. This includes considering the hidden and output units as maximally active or inactive by choosing hard-limit activation, approximating continuous activation functions by a threshold function, or changing the activation function steepness to act on the slope. This assumption is used to reduce the search by considering the network links to carry signals equal to the weights or no signal at all. 

Various heuristics have been employed to simplify the search process. This includes limiting the search depth by setting a maximum number of rule antecedents or considering specific combinations of rule antecedents that occur in the dataset. Some studies have also proposed heuristic search methods to find the best rules, such as CN2 \cite{clark1989}, or reinforcement learning \cite{ribeiro2018}. Stochastic search techniques, guided by Genetic Algorithms, have been used to discover effective rules \cite{johansson2004, lu2006, yedjour2018, yedjour2020}.

Alternative approaches involve searching for rules using association rule-based algorithms, with the Apriori algorithm being prominently used \cite{lu1995,setiono1995,setiono1996,lu1996,setiono1997b,hruschka1998,Hruschka2006,kaikhah2006}.

\begin{table*}[h]%
\centering
\caption{Summary of the explore \& test methods for rule extraction from feedforward neural networks}\label{tab:explore_test}
\begin{tabular*}{\textwidth}{@{\extracolsep\fill}llllll@{\extracolsep\fill}}
\hline
Translucency & Design & Application & Reference & Tool name & Technique \\
\hline
\multirow[t]{10}{*}{Pedagogical} & \multirow[t]{10}{*}{Post\textendash hoc} & A & \cite{saito1988}& \textendash & Heuristic search \\
\hline
& & A &\cite{saito1990}& RN & Sochastic search \\
& & A &\cite{craven1994} & \textendash & Stochastic search \\
& & A &\cite{zhou2000} & STARE & Heurostic search  \\
& & A &\cite{milare2001} & \textendash & heuristic search \\
& & A &\cite{johansson2004} & G-REX & Stochastic search \\
& & A &\cite{lofstrom2004} & CaST & heuristic search \\
& & A &\cite{yedjour2018} & NNGR & Stochastic search \\
& & A &\cite{ribeiro2018} & Anchors & Heuristic search \\
& & A &\cite{yedjour2020} & MNNGR & Stochastic search \\
\hline
\multirow[t]{21}{*}{Decompositional} & \multirow[t]{10}{*}{Post\textendash hoc} & S & \cite{gallant1988} & MACIE & Subset\textendash type \\
& & S & \cite{fu1991, fu1994} & KT\textendash algorithm & Subset\textendash type\\
& & S & \cite{towell1993} & Subset & Subset\textendash type\\
& & S & \cite{sethi1994} & \textendash & Subset\textendash type\\
& & S & \cite{hayward1996} & LAP & Subset\textendash type\\
& & S & \cite{taha1999} & Partal\textendash RE & Subset\textendash type\\
& & S & \cite{krishnan1999} & COMBO & Subset\textendash type\\
& & S & \cite{kim2000} & OAS & Subset\textendash type\\
& & S & \cite{garcez2001} & \textendash & Subset\textendash type\\
\cline{2-6}
& \multirow[t]{11}{*}{Intrinsic} & S & \cite{mcmillan1991a, mcmillan1991b} & \textendash & Subset\textendash type\\
& & S & \cite{towell1993, craven1993a, craven1993b} & MofN & Subset\textendash type\\
& & S & \cite{blasig1993} & GDS & Subset\textendash type \\
& & S & \cite{alexander1994} & \textendash & Subset\textendash type \\
& & S & \cite{lu1995,setiono1995,setiono1996} & NeuroRule & Association rules \\
& & S & \cite{lu1996,setiono1997b} & RX & Association rules \\
& & S & \cite{decloedt1996} & RULE\textendash OUT & Subset\textendash type \\
& & S & \cite{arbatli1997} & \textendash & Subset\textendash type  \\
& & S & \cite{hruschka1998} & Modified RX & Association rules \\
& & S & \cite{Hruschka2006} & \textendash & Association rules \\
& & S & \cite{lee2021} & OAS & Subset\textendash type \\
\hline
\multirow[t]{6}{*}{Eclectic} & \multirow[t]{3}{*}{Post\textendash hoc} & S & \cite{thrun1993,thrun1994, palade2001} & VIA & Heuristic and stochastic search \\ 
& & S & \cite{garcez2001} & \textendash  & Exhaustive search\\
& & S & \cite{craven1994} & \textendash & Stochastic search \\
\cline{2-6}
& \multirow[t]{3}{*}{Intrinsic} & S & \cite{duch1998} & C\textendash MLP2LN & Heuristic search \\
& & S & \cite{lu2006} & \textendash & Stochastic search \\
& & S & \cite{kaikhah2006} & \textendash & Association rules \\
\hline
\end{tabular*}
\end{table*}

The process of validating the extracted rules includes two main approaches within the context of explore \& test methods. The first approach involves assessing whether the extracted rules effectively activate the network units. The second approach, proposed by Thrun \cite{thrun1993}, is known as the Validity-Interval-Analysis (VIA) algorithm. This method ensures the consistency of the generated rules by propagating the rule constraints throughout the network. It employs linear programming techniques to determine the extent to which the input constraints propagate consistently through the activations of the network. 

The explore \& test methods, which involve extracting rules from the network after training, can pose challenges, particularly for large networks. Intrinsic methods are thus employed to mitigate these challenges by simplifying the network structure, making rule extraction more feasible. One approach to achieving this is by imposing regularization constraints on the network parameters or activation values during training. These regularization constraints help restrict the network's complexity, resulting in a sparser dependency structure. As a result, mapping the network to a set of rules becomes more straightforward, facilitating the extraction of rules from the trained network.

\subsection{Induced models based methods}\label{sec4-2}

Proxy models for neural networks can also be constructed using alternative techniques, such as decision trees, from which rules can be extracted. This section reviews studies on proxy models that are capable of providing explanations in the form of rules. These methods can be categorized as either local or global, depending on whether they provide explanations for a specific example or the entire behavior of the network. These methods are summarised in Table \ref{tab:induced_models_methods}.

\subsubsection{Local methods}

Using the following procedure, local methods create a proxy model to generate explanations of the network decision for a particular input example. Given an input $x$ and its network prediction:

\begin{itemize}
    \item Generate a neighborhood of similar samples of $x$.
    \item Calculate the predictions of the network of the generated neighborhood samples.
    \item Train an interpretable proxy model on the generated samples as inputs and the network predictions as targets to mimic its behavior at that specific input area.
    \item Extract rules from the proxy model.
\end{itemize}

Local proxies include program expression trees \cite{singh2016} and Local Rule Based Explanations (LORE) \cite{guidotti2018}. The program expression trees method uses programs as interpretable representations to explain local decisions. These programs are similar to the pseudo codes used to represent decision rules. The neighborhood is generated using perturbation techniques, and program induction is performed using a Simulated Annealing algorithm. The LORE method derives local decision rules and counterfactuals to characterize the network behavior locally. They use Genetic algorithms to generate the neighborhood and a decision tree local surrogate to approximate the network. 

\subsubsection{Global methods}

Global methods build proxy models to mimic the behavior of the network and explain it globally. Rules can be extracted following pedagogical, decompositional, or eclectic approaches according to how the proxy model is fitted.

\textbf{Pedagogical methods} include decision trees-based proxy models considering the network as a black box and using input data and the network's predictions as targets to learn the proxies. The standard approach considers as targets the hard labels given by the network, i.e., discrete classes \cite{craven1995, schmitz1999, milare2001}. Distillation methods, however, consider soft labels as targets, which are the probabilities predicted by the network \cite{che2016, frosst2017,liu2018}. Another approach consists of using decision tables to approximate the network \cite{sethi2012} and extract rules. Boolean functions have also been used to approximate the network behavior by minimizing truth tables Boolean expressions \cite{taha1999, mereani2019}.

\textbf{Decompositional methods} use surrogates to approximate each unit in the network. In this case, the training data are the unit inputs, and the targets are the unit activation values. The decompositional approach can be post-hoc or intrinsic. In the post-hoc case, this approach does not impose any particular training procedure and is applied after the network training. The intrinsic case, however, applies particular training procedures to simplify the rule extraction task. \cite{tsukimoto2000} proposed an approximation of the network’s units using Boolean functions. The Continuous/discrete Rule Extractor via Decision tree induction (CRED) \cite{sato2001} consists of approximating the hidden and output units using decision trees. \cite{zilke2016} extended it to the case of networks with multiple layers using the Deep neural network Rule Extraction via Decision tree induction (DeepRED) technique and pruning unimportant inputs. The Rule Extraction Methodology from DNN (REM-D) \cite{shams2021} improves the DeepRED by using decision trees C5 instead of C4.5 as a proxy model and merging the extracted rules incrementally as soon as they are generated. In DeepRED, rules are extracted for all layers and merged, improving time and memory consumption. \cite{zarlenga2021} proposed a method to overcome the exponential runtime with respect to the training set and the network depth encountered in DeepRED and REM-D. These methods are post-hoc, except for the DeepRED, which considers a pruning stage.

Other intrinsic decompositional methods were proposed and follow mainly this procedure:
\begin{itemize}
    \item Train a network using regularization techniques.
    \item Prune the network to remove insignificant units.
    \item Fit a surrogate model at each unit level and extract rules.
    \item Derive and simplify rules expressing the relationship between the inputs and the outputs.
\end{itemize}

This category includes \cite{setiono2002, setiono2004} studies that simplify the network using regularization and pruning techniques and approximates the hidden and output units using piecewise linear functions. These functions are used to express oblique rules.

\textbf{Eclectic methods} consider fitting proxy models while integrating some knowledge from the network internals. \cite{duch1998} used regularization techniques to push the network to behave like a Boolean function. \cite{setiono2008} considered regularizing and pruning the network, including the inputs, and extracting rules using a decision tree proxy. \cite{thiagarajan2016} proposed an approach where the network decision is explained in terms of the latent variables instead of the input features. Each input example is mapped to feature factors from the representation space for which explanations are generated from a decision tree-based proxy method. \cite{bondarenko2017} incorporated knowledge about the network's last hidden layer into the decision tree extraction process. \cite{wu2018} applied a particular regularization technique incorporating  constraint into the loss function to make the network decision boundaries behave like a decision tree. The fitted decision tree proxy gives better results than using standard regularization techniques.

\begin{table*}[!t]
\centering
\caption{Summary of the induced models based methods for rule extraction from feedforward neural networks}\label{tab:induced_models_methods}
\footnotesize
\begin{tabular*}{\textwidth}{@{\extracolsep\fill}lllllll@{\extracolsep\fill}}
\hline
Translucency & Design & Application & Scope & Reference & Tool name & Technique \\
\hline
\multirow[t]{8}{*}{Pedagogical} & \multirow[t]{8}{*}{Post\textendash hoc} &  A & G & \cite{craven1995}& TREPAN & Decision trees \\
\hline
& & A & G &\cite{schmitz1999}& ANN-DT & Decision trees \\
& & A & G & \cite{taha1999} & BIO-RE & Boolean functions \\
& & A & G & \cite{milare2001} & - & Decision trees \\
& & A & G & \cite{sethi2012} & KDRuleEx & Decision tables \\
& & A & L &\cite{singh2016} & - & Program learner \\
& & A & G &\cite{che2016} & - & Decision trees \\
& & A & G &\cite{frosst2017} & Distilling  & Decision trees \\
& & A & L &\cite{guidotti2018} & LORE & Decision trees \\
& & A & G &\cite{liu2018} & - & Decision trees \\
& & A & G &\cite{liu2018} & - & Boolean functions \\
\hline
\multirow[t]{2}{*}{Decompositional} & \multirow[t]{2}{*}{Post\textendash hoc} & S & G & \cite{tsukimoto2000} & - & Boolean functions \\
& & S & G & \cite{sato2001} & CRED & Decision trees \\
& & S & G & \cite{shams2021} & REM-D & Decision trees\\
& & S & G & \cite{zarlenga2021} & ECLAIRE & Decision trees \\
\cline{2-7}
& \multirow[t]{2}{*}{Intrinsic} &  S & G & \cite{setiono2002} & REFANN & Piecewise linear functions \\
& & S & G & \cite{setiono2004} & REFANN & Piecewise linear functions \\
& & S & G & \cite{zilke2016} & DeepRED & Decision trees \\
\hline
\multirow[t]{2}{*}{Eclectic} & \multirow[t]{2}{*}{Post\textendash hoc} & S & G & \cite{thiagarajan2016} & Tree-View & Decision Trees \\ 
\cline{2-7}
& \multirow[t]{2}{*}{Intrinsic} & S & G & \cite{duch1998} & MLP2LN & Boolean functions \\
& & S & G & \cite{setiono2008} & RE-RX & Decision trees \\
& & S & G & \cite{bondarenko2017} & NNKX & Decision Trees \\
& & S & G & \cite{wu2018} & Tree regularization & Decision trees \\
\hline
\end{tabular*}
\end{table*}


\subsection{Attribution methods}\label{sec4-3}

Attribution-based methods provide explanations by defining the input features' effect on the outputs. They have recently gained importance as a class of interpretability methods for DL models. To simplify the challenge of explaining a deep network, one can identify the important features, i.e., inputs or hidden units, for a particular output. The techniques for performing such a task can have different appellations, such as saliency, relevance, importance, or attribution. Feature importance can be visualized using heat maps or threshold masks. Many approaches are based on back-propagation of the signal output to input or local approximation of the decision boundaries using liner classifiers or decision trees. The existing methods can be classified into \cite{townsend2019}:

\begin{itemize}
    \item Backpropagation of feature importance: these methods show positive contributions to a class. They are based on the backpropagation of signals from a deeper layer to a shallower one. Usually performed from the output layer to the input layer. The signal could be the activation, gradient, or some other metric derived from the output (e.g. deconvolution, gradient-based methods, and guided-backpropagation).
    \item Backpropagation of bipolar feature importance: they can indicate evidence for and against a class (e.g. layerwise relevance propagation and DeepLIFT). 
    \item Perturbation methods: are based on evaluating the effect of removing or modifying a feature on the prediction (e.g. Prediction Difference Analysis, LIME, and SHAP). 
\end{itemize}
	 
In recent literature reviews, there has been a growing interest in attribution-based methods for explaining DL models. These methods primarily focus on providing local explanations, which involve explaining a specific decision made by the model in relation to a single input by highlighting the important features that contributed to that decision. \cite{ribeiro2016} proposed one approach towards achieving global interpretation from local attribution-based explanations. Their method involves explaining a set of representative individual instances and formulating the problem of defining the optimal set of instances as a submodular optimization problem. This approach aims to provide a holistic understanding of the model's decision-making process by considering multiple instances together rather than focusing solely on individual inputs. It has also been observed that most of attribution-based methods do not establish the underlying relationships between the extracted important features, for instance, in the form of logical rules \cite{townsend2019}. 

The idea of defining the most important features and extracting important relationships between them in the context of feedforward neural networks is not new. As early as the 1990s, this challenge was addressed from different perspectives. For instance, \cite{baba1990} proposed a causal index reflecting the influence of an individual input $x_i$ on a specific output $o_j$. In the network, different paths could link an input $x_i$ to an output $o_j$, with each path defined as a succession of links. The causal index $CI_{ij}$ is calculated as the sum of the products of path weights along all the paths between $x_i$ and $o_j$ in the network. The sign of the index reflects a positive or a negative correlation between the input and the output signals. The absolute value of the index gives the degree of correlation. \cite{goh1993} proposed a sensitivity-based approach to calculating features' importance globally. The Point Sensitivity Index (PSI) is calculated for the entire range of the input showing decisive inputs with peaks in their PSI at critical values. The Sensitivity Causal Index (SCI) is an indicator of the global impact of a feature on an output. Unlike \cite{baba1990} method using network weights, sensitivity analysis is applied without using the network's internals. \cite{tickle1994} proposed the DEcision DEteCtion by rule extraction from neural networks (DEDEC) method to rank the input features that should be considered for rule extraction. DEDEC analyses the weight vectors to determine causal factors and functional dependencies in a trained neural network. Causal Attributions \cite{chattopadhyay2019} consider the neural network as a Structural Causal Model to calculate the global causal effect of each feature on the output using the Average Causal Effect (ACE) \cite{pearl2009}. Based on the magnitude and the sign of the feature importance indicator, rules between the inputs and the outputs can be extracted in the form ``IF $x_2$ is large THEN $o_1$ is large”. For simple data cases, the decision boundaries can be clearly separated \cite{chattopadhyay2019}.

The Building Representations for AI using Neural NEtworks (BRAINNE) approach \cite{sestito1990, sestito1991, sestito1993, bloomer1996} analyzes the network weights to define the important inputs to use for rule extraction. First, the method was proposed for single-layered networks composed of input and output layers trained with Hebb’s rule \cite{chakraverty2019}. The method was then extended to the network with one hidden layer trained using Hebb’s rule and backpropagation, with a particular architecture considering the network outputs as inputs. \cite{bloomer1996} extended the method to extract disjunctive rules using a hybrid network composed of an unsupervised Kohonen model \cite{kohonen1990} and a second supervised part of a standard output layer. \cite{gupta1999} proposed the Generalized Analytic Rule Extraction (GLARE), using an analytical approach to rank the inputs’ importance for each output class based on the network weights magnitude. \cite{augasta2012} and \cite{biswas2017, biswas2018} used a reverse engineering technique to identify the important inputs and find mandatory input data ranges to classify a particular output class. 

Table \ref{tab:attribution_methods} summarizes the studies based on attribution methods. \cite{bloomer1996} method is considered decompositional as they extract rules at an intermediate level between inputs and outputs layers and proceed with substitution to get rules linking inputs and output directly. It is also intrinsic as it considers a hybrid network using a self-organizing map to cluster the inputs and a standard feedforward network to classify them. \cite{baba1990}, \cite{sestito1990, sestito1991, sestito1993}, \cite{tickle1994}, and \cite{gupta1999} used weights to define features’ importance, hence their methods can be classified as eclectic. \cite{goh1993},  \cite{augasta2012}, \cite{biswas2017, biswas2018}, and \cite{chattopadhyay2019} are pedagogical as they do not use the model’s internals. However, \cite{chattopadhyay2019} considers the network as a Structural Causal Model to use causality principles and calculate Causal Attributions. Hence it is considered a model-specific method. All attribution methods are global.

\begin{table*}[!t]
\centering
\caption{Summary of the attribution methods for rule extraction from feedforward neural networks}\label{tab:attribution_methods}
\begin{tabular*}{\textwidth}
{{p{2.4cm}p{1.3cm}p{1.3cm}p{1.5cm}p{4cm}p{3cm}}}
\hline
Translucency & Design & Application & Reference & Tool name & Technique \\
\hline
\multirow[t]{3}{*}{Pedagogical} & \multirow[t]{3}{*}{Post\textendash hoc} & A & \cite{goh1993} & Point Sensitivity Index (PSI) and Sensitivity Causal Index (SCI) & Causality and sensitivity  \\ 
& & A & \cite{augasta2012, biswas2017} & RxREN & Reverse engineering  \\
& & A & \cite{biswas2018} & RxNCM & Reverse engineering  \\ 
\hline
\multirow[t]{1}{*}{Decompositional} & \multirow[t]{1}{*}{Intrinsic} & S & \cite{bloomer1996} & BRAINNE & Weights analysis \\
\hline
\multirow[t]{5}{*}{Eclectic} & \multirow[t]{5}{*}{Post\textendash hoc} & S & \cite{baba1990} & Causal Index & Gradient \\ 
& & S & \cite{sestito1990, sestito1991, sestito1993} & BRAINNE & Network weights analysis \\
& & S & \cite{tickle1994} & DEDEC & Network weights analysis \\
& & S & \cite{gupta1999} & GLARE & Network weights analysis \\
& & S & \cite{chattopadhyay2019} & Causal Attributions & Causality (Average Causal Effect) \\ 
\hline
\end{tabular*}
\end{table*}


\subsection{Optimization methods}\label{sec4-4}

The rule extraction task from neural networks can be formulated as an optimization problem. The objective is to find the inputs’ optimal values that maximize a particular class's probability. 

Considering a network with one hidden layer and sigmoid activations, the $j^{th}$ output activation can be expressed by Equation (\ref{eq:optimization_method}). It is a non-linear function of the inputs parameterized by the network weights $W_1$ and $W_2$, where $W_1$ are the weights linking the inputs to the hidden units and $W_2$ are the weights linking the hidden units to the output. After network training, the weights are fixed, and the activation function of the $j^{th}$ output can be expressed as a function of the inputs. Finding the input vector which maximizes $Y_j(x_i)$ is an optimization problem. If the decision variables or inputs $x_i$ are binary and the objective function $Y_j$ is nonlinear, the problem can be expressed as a nonlinear integer programming problem. Solving this problem enables expressing rules linking the input features to the output class. Table \ref{tab:optimization_methods} summarizes studies on rule extraction from neural networks based on optimization techniques. All methods are specific to neural networks and provide global explanations. \cite{ozbakir2009,ozbakir2010} solved the optimization problem using the Touring Ant Colony Optimization (TACO) algorithm, \cite{kahramanli2009} using Artificial Immune Systems algorithm, and \cite{uzun2016} using Variable Neighbourhood Search algorithm. \cite{ozbakir2010} used the TACO method with an intrinsic approach where the rules are extracted during network training, and their quality metrics are used to regularize the network towards generating good rules. 

\begin{equation}\label{eq:optimization_method}
Y_j = \frac{1}{1+e^{-[{\sum_{j=1}^{J} (W_2)_{j,k}(\frac{1}{1+e^-[\sum_{i=1}^{I}x_i(W_1)_{}i,j]})}]}} 
\end{equation}

$x_i = 0 \: or \:  1$

Optimization-based techniques for extracting rules from neural networks are not suitable for deep networks. This approach faces challenges in scaling up to deep networks due to the increased complexity that arises with additional hidden layers. Furthermore, applying it to high-dimensional data can be challenging. Additionally, accessing the weights of the network is required.

\begin{table*}[t]
\centering
\caption{Summary of the optimization methods for rule extraction from feedforward neural networks}\label{tab:optimization_methods}
\begin{tabular*}{\textwidth}{@{\extracolsep\fill}llllll@{\extracolsep\fill}}
\hline
Translucency & Design & Reference & Tool name & Technique \\
\hline
\multirow[t]{2}{*}{Eclectic} & \multirow[t]{2}{*}{Pos\textendash thoc} & \cite{ozbakir2008,ozbakir2009} & TACO & Ant Colony Algorithm  \\ 
& & \cite{kahramanli2009} & - & Artificial Immune Systems algorithm \\
& & \cite{uzun2016} & - & Variable Neighbourhood Search  \\
\cline{2-5}
& \multirow[t]{2}{*}{Intrinsic} & \cite{ozbakir2010} & DIFACONN-miner & Ant Colony Algorithm \\
\hline
\end{tabular*}
\end{table*}


\subsection{Hybrid methods}\label{sec4-5}

Hybrid methods combine different techniques from the previously presented for rules extraction from neural networks. Table \ref{tab:hybrid_methods} presents hybrid methods, which are all global and specific. Considering networks with one hidden layer, one approach expresses the relationship between inputs and hidden units as linear inequalities. Rules between hidden and output units can be extracted using an explore \& test method \cite{setiono1997a,odajima2006}, Boolean functions \cite{bologna1998}, or decision trees \cite{bologna2000, setiono2000}. The rule extraction problem can also be expressed as a linear programming standard problem to extract rules between binary or discretized input features and hidden units \cite{taha1999}. 

\cite{saito2002} proposed the Rule extraction from Neural networks version 2 (RN2) to produce regression rules on a specific network architecture to allow extraction of rules with logical formula on the premise part over the nominal attributes and a polynomial equation over the continuous attributes in the conclusion part. They used decision trees to generate rules between nominal attributes and discretized hidden activations. The rules conclusions are obtained by injecting the activation values in the polynomial term, including the continuous values. \cite{chakraborty2019} proposed the Eclectic Rule Extraction from Neural Network Recursively (ERENNR) method using a rule learner to extract rules between hidden and output units. The conditions are then replaced by input ranges obtained from analyzing each hidden unit from its classified patterns. It was extended to Eclectic Rule Extraction from Neural Network with Multi Hidden Layer (ERENN MHL) by \cite{chakraborty2020} method for multi-hidden layer networks. 

\begin{table*}
\centering
\caption{Summary of hybrid methods for rule extraction from feedforward neural networks}\label{tab:hybrid_methods}
\begin{tabular*}{\textwidth}{@{\extracolsep\fill}lllll@{\extracolsep\fill}}
\hline
Translucency & Design & Reference & Tool name & Technique \\
\hline
\multirow[t]{2}{*}{Decompositional} & \multirow[t]{2}{*}{Post\textendash hoc} & \cite{taha1999} & Full-RE & Optimization and explore \& test\\ 
& & \cite{saito2002} & RN2 & Decision trees and analytical expressions\\ 
\cline{2-5}
& \multirow[t]{2}{*}{Intrinsic} & \cite{setiono1997a} & Neurolinear &  linear inequalities and explore \& test\\
& & \cite{setiono2000} & FERNN &  linear inequalities and decision trees\\
& & \cite{odajima2006} & GRG & Linear inequalities and explore \& test \\
& & \cite{huynh2009} & \textendash & Decision trees and linear inequalities\\
& & \cite{chakraborty2020} & ERENN MHL &  Attribution methods and Boolean functions \\
\hline
\multirow[t]{2}{*}{Eclectic} & \multirow[t]{2}{*}{Post\textendash hoc} & \cite{bologna1998} & IMLP & Linear inequalities and Boolean functions \\ 
& & \cite{bologna2000} & DIMLP & Linear inequalities and decision trees \\
\cline{2-5}
& \multirow[t]{2}{*}{Intrinsic} & \cite{chakraborty2019} & \textendash & Attribution methods and Boolean functions\\
\hline
\end{tabular*}
\end{table*}

\section{Discussion}\label{sec5} 

Most of the rule extraction methods from neural networks reviewed in this study were proposed for shallow networks with a depth not exceeding three hidden layers \cite{he2020}. Rule extraction from neural networks can be computationally expensive, especially for larger, more complex networks or datasets. The complexity of a rule extraction method depends on various factors, such as the type of extraction algorithm, the architecture and the size of the network, and the complexity of the data being processed. The explore \& test, optimization, and decompositional-based extraction methods particularly involve complex operations even with shallow networks. Moreover, deep neural networks with a large number of layers and nodes may require more computational resources and time to extract rules. Furthermore, when dealing with high-dimensional data or intricate patterns, the level of complexity increases.

Decompositional-based techniques represent an important part of rule extraction methods from neural networks. The decompositional approach introduces intermediate terms corresponding to each of the hidden and output units in the network. The intuition behind this is that rules extracted at individual units can be merged to explain the network behavior. This approach, however, assumes that each hidden and output unit can behave like a Boolean variable, and the individual hidden units of the network correspond to meaningful concepts in the context of the domain. Relying on intuition-based methods for interpreting deep neural networks can be misleading, and it is important to determine if single-unit behavior can help understanding deep neural networks. Decompositional methods focus on the properties of individual units by finding rules that activate a specific unit. Neural networks, however, learn distributed representations in which the activations of many hidden units encode each concept, and each hidden unit plays a part in representing many different concepts. Given such representations, the right intermediate terms might represent patterns of activity across groups of hidden units instead of individual hidden units themselves. 

For a rule extraction method to be adaptable even for large networks, the following properties defined by \cite{thrun1993} are highly desirable: (i) it should have no architectural requirements and is able to work with all networks; (ii) it has no training requirements and makes no assumptions about how the network was built and trained, (iii) it provides correct rules describing the underlying network, (iv) it has high expressive power in terms of the language and compactness of the rules. Given these properties, post-hoc model agnostic proxy methods are the most appropriate to use with deep neural networks. Proxy methods are currently a popular approach for interpreting DL models. One of their key advantages is their agnostic nature, as they treat the black-box model as an oracle and only require a dataset along with the corresponding model's predictions. They offer a simple way to interpret. However, it's important to note that proxy methods have their limitations and may not always provide a complete or accurate explanation of the network behavior, as they are approximations and not a direct representation of the original model. Attribution methods also offer a cheap way to local interpretation of DL models. This helps in understanding why a particular prediction was made in a specific context, which is important in object detection, image segmentation, or natural language processing tasks. However, attribution methods might struggle to capture the behavior of the entire model to provide concise global explanations. 

Finally, the extraction of interpretable rules from neural networks includes an essential phase that often remains underemphasized in the existing literature: the pruning of extracted rules. This phase becomes particularly significant when employing decompositional approaches or dealing with large networks, as the extraction process can yield rule sets with an excessive number of redundant or irrelevant rules. Such redundancy and irrelevance not only affect the interpretability but also limit the generalization performance. Consequently, the development of efficient techniques for rule pruning becomes imperative, enabling the selective removal of unnecessary rules while retaining the crucial information encoded in the network. Effective rule pruning not only enhances the interpretability of neural networks but also enhances their computational efficiency and scalability in rule extraction methodologies. By facilitating the extraction of simpler and more concise rule sets, these pruning techniques enable a comprehensive understanding of the underlying behavior of the neural network. Consequently, investing in the advancement of efficient rule pruning techniques \cite{berrado2007,azmi2019} is important in interpreting complex neural networks through simple decision rules.

\section{Conclusion}\label{sec6}

Improving the interpretability of ML models is crucial for the successful deployment of AI enabled systems. This paper aimed to contribute to this objective by focusing on a specific tool for interpretability, namely rule extraction for neural networks. The main objective of this research was to conduct a systematic literature review to identify and understand the various approaches used to generate rules from feedforward neural networks.

The findings of this study indicate that the field of rule extraction from neural networks has been actively explored for over two decades, resulting in the development of diverse methods. While most of the proposed techniques are well-suited for shallow networks, it was observed that some of them can also be adapted for DL models.

Rules, due to their simplicity and transparency, offer an intuitive means of explaining neural network behavior to users. Therefore, this study provides a comprehensive introduction for researchers interested in this field, laying the groundwork for further investigations and advancements.

It is important to note that this study specifically focused on feedforward networks with supervised learning tasks and crisp extracted rules. However, there is ample opportunity for future research to extend these findings to rule extraction from other types of neural networks, such as RNN and CNN models. Additionally, it would be valuable to explore rule extraction methods in the context of other ML techniques and to include approaches related to fuzzy rules extraction.

In conclusion, this research contributes to the understanding of rule extraction methods for feedforward neural networks and serves as a valuable resource for researchers aiming to enhance interpretability in AI systems. Expanding this work to encompass other neural network types and ML methods, as well as incorporating fuzzy rule extraction, presents exciting avenues for future exploration in this field.

\bibliographystyle{abbrvnat}


\begin{thebibliography}{110}
\providecommand{\natexlab}[1]{#1}
\providecommand{\url}[1]{\texttt{#1}}
\expandafter\ifx\csname urlstyle\endcsname\relax
  \providecommand{\doi}[1]{doi: #1}\else
  \providecommand{\doi}{doi: \begingroup \urlstyle{rm}\Url}\fi

\bibitem[Alexander and Mozer(1994)]{alexander1994}
J.~Alexander and M.~C. Mozer.
\newblock Template-based algorithms for connectionist rule extraction.
\newblock In G.~Tesauro, D.~Touretzky, and T.~Leen, editors, \emph{Advances in Neural Information Processing Systems}, volume~7, pages 609--616. MIT Press, 1994.
\newblock URL \url{https://proceedings.neurips.cc/paper/1994/file/24896ee4c6526356cc127852413ea3b4-Paper.pdf}.

\bibitem[Andrews et~al.(1995)Andrews, Diederich, and Tickle]{andrews1995}
R.~Andrews, J.~Diederich, and A.~B. Tickle.
\newblock Survey and critique of techniques for extracting rules from trained artificial neural networks.
\newblock \emph{Knowledge-based systems}, 8\penalty0 (6):\penalty0 373--389, 1995.

\bibitem[Augasta and Kathirvalavakumar(2012)]{augasta2012}
M.~G. Augasta and T.~Kathirvalavakumar.
\newblock Reverse engineering the neural networks for rule extraction in classification problems.
\newblock \emph{Neural processing letters}, 35:\penalty0 131--150, 2012.
\newblock \doi{10.1007/s11063-011-9207-8}.

\bibitem[Azmi et~al.(2019)Azmi, Runger, and Berrado]{azmi2019}
M.~Azmi, G.~C. Runger, and A.~Berrado.
\newblock Interpretable regularized class association rules algorithm for classification in a categorical data space.
\newblock \emph{Information Sciences}, 483:\penalty0 313--331, 2019.

\bibitem[Baba et~al.(1990)Baba, Enbutu, and Yoda]{baba1990}
K.~Baba, I.~Enbutu, and M.~Yoda.
\newblock Explicit representation of knowledge acquired from plant historical data using neural network.
\newblock In \emph{1990 IJCNN International Joint Conference on Neural Networks}, volume~3, pages 155--160, 1990.
\newblock \doi{10.1109/IJCNN.1990.137838}.

\bibitem[Beck et~al.(2016)Beck, Koch, and Weiskopf]{beck2016}
F.~Beck, S.~Koch, and D.~Weiskopf.
\newblock Visual analysis and dissemination of scientific literature collections with survis.
\newblock \emph{IEEE Transactions on Visualization and Computer Graphics}, 22\penalty0 (1):\penalty0 180--189, 2016.
\newblock \doi{10.1109/TVCG.2015.2467757}.

\bibitem[Berrado and Runger(2007)]{berrado2007}
A.~Berrado and G.~C. Runger.
\newblock Using metarules to organize and group discovered association rules.
\newblock \emph{Data mining and knowledge discovery}, 14:\penalty0 409--431, 2007.

\bibitem[Biswas et~al.(2017)Biswas, Chakraborty, Purkayastha, Roy, and Thounaojam]{biswas2017}
S.~K. Biswas, M.~Chakraborty, B.~Purkayastha, P.~Roy, and D.~M. Thounaojam.
\newblock Rule extraction from training data using neural network.
\newblock \emph{International Journal on Artificial Intelligence Tools}, 26\penalty0 (03):\penalty0 1750006, 2017.
\newblock \doi{10.1142/S0218213017500063}.
\newblock URL \url{https://doi.org/10.1142/S0218213017500063}.

\bibitem[Biswas et~al.(2018)Biswas, Chakraborty, and Purkayastha]{biswas2018}
S.~K. Biswas, M.~Chakraborty, and B.~Purkayastha.
\newblock A rule generation algorithm from neural network using classified and misclassified data.
\newblock \emph{International Journal of Bio-Inspired Computation}, 11\penalty0 (1):\penalty0 60--70, 2018.
\newblock \doi{10.1504/IJBIC.2018.090070}.
\newblock URL \url{https://www.inderscienceonline.com/doi/abs/10.1504/IJBIC.2018.090070}.

\bibitem[Blasig(1993)]{blasig1993}
R.~Blasig.
\newblock Gds: Gradient descent generation of symbolic classification rules.
\newblock In J.~Cowan, G.~Tesauro, and J.~Alspector, editors, \emph{Advances in Neural Information Processing Systems}, volume~6, pages 1993--1100. Morgan-Kaufmann, 1993.
\newblock URL \url{https://proceedings.neurips.cc/paper/1993/file/4e0928de075538c593fbdabb0c5ef2c3-Paper.pdf}.

\bibitem[Bloomer et~al.(1996)Bloomer, Dillon, and Witten]{bloomer1996}
W.~Bloomer, T.~Dillon, and M.~Witten.
\newblock Hybrid brainne: a method for developing symbolic disjunctive rules from a hybrid neural network.
\newblock In \emph{1996 IEEE International Conference on Systems, Man and Cybernetics. Information Intelligence and Systems (Cat. No.96CH35929)}, volume~4, pages 2745--2750, 1996.
\newblock \doi{10.1109/ICSMC.1996.561374}.

\bibitem[Bologna(2000)]{bologna2000}
G.~Bologna.
\newblock Symbolic rule extraction from the dimlp neural network.
\newblock In S.~Wermter and R.~Sun, editors, \emph{Hybrid Neural Systems}, pages 240--254, Berlin, Heidelberg, 2000. Springer Berlin Heidelberg.
\newblock ISBN 978-3-540-46417-4.

\bibitem[Bologna and Pellegrini(1998)]{bologna1998}
G.~Bologna and C.~Pellegrini.
\newblock Constraining the mlp power of expression to facilitate symbolic rule extraction.
\newblock In \emph{1998 IEEE International Joint Conference on Neural Networks Proceedings. IEEE World Congress on Computational Intelligence (Cat. No.98CH36227)}, volume~1, pages 146--151, 1998.
\newblock \doi{10.1109/IJCNN.1998.682252}.

\bibitem[Bondarenko et~al.(2017)Bondarenko, Aleksejeva, Jumutc, and Borisov]{bondarenko2017}
A.~Bondarenko, L.~Aleksejeva, V.~Jumutc, and A.~Borisov.
\newblock Classification tree extraction from trained artificial neural networks.
\newblock \emph{Procedia Computer Science}, 104:\penalty0 556--563, 2017.
\newblock ISSN 1877-0509.
\newblock \doi{10.1016/j.procs.2017.01.172}.
\newblock URL \url{https://www.sciencedirect.com/science/article/pii/S1877050917301734}.
\newblock ICTE 2016, Riga Technical University, Latvia.

\bibitem[Chakraborty et~al.(2019)Chakraborty, Biswas, and Purkayastha]{chakraborty2019}
M.~Chakraborty, S.~K. Biswas, and B.~Purkayastha.
\newblock Rule extraction from neural network using input data ranges recursively.
\newblock \emph{New Generation Computing}, 37:\penalty0 67--96, 2019.
\newblock \doi{10.1007/s00354-018-0048-0}.

\bibitem[Chakraborty et~al.(2020)Chakraborty, Biswas, and Purkayastha]{chakraborty2020}
M.~Chakraborty, S.~K. Biswas, and B.~Purkayastha.
\newblock Rule extraction from neural network trained using deep belief network and back propagation.
\newblock \emph{Knowledge and Information Systems}, 62:\penalty0 3753--3781, 2020.
\newblock \doi{10.1007/s10115-020-01473-0}.

\bibitem[Chakraverty et~al.(2019)Chakraverty, Sahoo, Mahato, Chakraverty, Sahoo, and Mahato]{chakraverty2019}
S.~Chakraverty, D.~M. Sahoo, N.~R. Mahato, S.~Chakraverty, D.~M. Sahoo, and N.~R. Mahato.
\newblock Hebbian learning rule.
\newblock \emph{Concepts of Soft Computing: Fuzzy and ANN with Programming}, pages 175--182, 2019.

\bibitem[Chattopadhyay et~al.(2019)Chattopadhyay, Manupriya, Sarkar, and Balasubramanian]{chattopadhyay2019}
A.~Chattopadhyay, P.~Manupriya, A.~Sarkar, and V.~N. Balasubramanian.
\newblock Neural network attributions: A causal perspective.
\newblock In K.~Chaudhuri and R.~Salakhutdinov, editors, \emph{Proceedings of the 36th International Conference on Machine Learning}, volume~97, pages 981--990. PMLR, 09--15 Jun 2019.
\newblock URL \url{https://proceedings.mlr.press/v97/chattopadhyay19a.html}.

\bibitem[Che et~al.(2016)Che, Purushotham, Khemani, and Liu]{che2016}
Z.~Che, S.~Purushotham, R.~Khemani, and Y.~Liu.
\newblock Interpretable deep models for icu outcome prediction.
\newblock In \emph{AMIA annual symposium proceedings}, volume 2016, pages 371--380, 2016.
\newblock URL \url{https://www.ncbi.nlm.nih.gov/pmc/articles/PMC5333206/}.

\bibitem[Clark and Niblett(1989)]{clark1989}
P.~Clark and T.~Niblett.
\newblock The cn2 induction algorithm.
\newblock \emph{Machine learning}, 3\penalty0 (4):\penalty0 261--283, 1989.

\bibitem[Craven and Shavlik(1995)]{craven1995}
M.~Craven and J.~Shavlik.
\newblock Extracting tree-structured representations of trained networks.
\newblock In D.~Touretzky, M.~Mozer, and M.~Hasselmo, editors, \emph{Advances in Neural Information Processing Systems}, volume~8, pages 24--30. MIT Press, 1995.
\newblock URL \url{https://proceedings.neurips.cc/paper/1995/file/45f31d16b1058d586fc3be7207b58053-Paper.pdf}.

\bibitem[Craven(1996)]{craven1996}
M.~W. Craven.
\newblock \emph{Extracting comprehensible models from trained neural networks}.
\newblock The University of Wisconsin-Madison, 1996.

\bibitem[Craven and Shavlik(1993{\natexlab{a}})]{craven1993a}
M.~W. Craven and J.~W. Shavlik.
\newblock Learning symbolic rules using artificial neural networks.
\newblock In \emph{Machine Learning Proceedings 1993}, pages 73--80. Morgan Kaufmann, San Francisco (CA), 1993{\natexlab{a}}.
\newblock ISBN 978-1-55860-307-3.
\newblock \doi{10.1016/B978-1-55860-307-3.50016-2}.
\newblock URL \url{https://www.sciencedirect.com/science/article/pii/B9781558603073500162}.

\bibitem[Craven and Shavlik(1993{\natexlab{b}})]{craven1993b}
M.~W. Craven and J.~W. Shavlik.
\newblock Understanding neural networks via rule extraction and pruning.
\newblock In \emph{Proceedings of the 1993 Connectionist Models Summer School}, page 184. Psychology Press, 1993{\natexlab{b}}.
\newblock URL \url{https://books.google.co.ma/books?hl=en&lr=&id=0SEBAwAAQBAJ&oi=fnd&pg=PA184&dq=Understanding+neural+networks+via+rule+extraction+and+pruning&ots=U5vxpNw3CH&sig=1eNzxACVFHmoEBk4M3fldMCKNZY&redir_esc=y#v=onepage&q=Understanding\%20neural\%20networks\%20via\%20rule\%20extraction\%20and\%20pruning&f=false}.

\bibitem[Craven and Shavlik(1994)]{craven1994}
M.~W. Craven and J.~W. Shavlik.
\newblock Using sampling and queries to extract rules from trained neural networks.
\newblock In W.~W. Cohen and H.~Hirsh, editors, \emph{Machine Learning Proceedings 1994}, pages 37--45. Morgan Kaufmann, San Francisco (CA), 1994.
\newblock ISBN 978-1-55860-335-6.
\newblock \doi{https://doi.org/10.1016/B978-1-55860-335-6.50013-1}.
\newblock URL \url{https://www.sciencedirect.com/science/article/pii/B9781558603356500131}.

\bibitem[{d'Avila Garcez} et~al.(2001){d'Avila Garcez}, Broda, and Gabbay]{garcez2001}
A.~S. {d'Avila Garcez}, K.~Broda, and D.~M. Gabbay.
\newblock Symbolic knowledge extraction from trained neural networks: A sound approach.
\newblock \emph{Artificial Intelligence}, 125\penalty0 (1):\penalty0 155--207, 2001.
\newblock ISSN 0004-3702.
\newblock \doi{10.1016/S0004-3702(00)00077-1}.
\newblock URL \url{https://www.sciencedirect.com/science/article/pii/S0004370200000771}.

\bibitem[Decloedt et~al.(1996)Decloedt, Os{\'o}rio, and Amy]{decloedt1996}
L.~Decloedt, F.~Os{\'o}rio, and B.~Amy.
\newblock Rule\_out method: A new approach for knowledge explicitation from trained ann.
\newblock In Andrews and Diederich, editors, \emph{Proceedings of the AISB’96 - Workshop on Rule Extraction from Trained Neural Nets}, pages 34--42, 1996.
\newblock URL \url{https://citeseerx.ist.psu.edu/viewdoc/download?doi=10.1.1.46.8091&rep=rep1&type=pdf}.

\bibitem[Duch et~al.(1998)Duch, Adamczak, and Gr{\k{a}}bczewski]{duch1998}
W.~Duch, R.~Adamczak, and K.~Gr{\k{a}}bczewski.
\newblock Extraction of logical rules from neural networks.
\newblock \emph{Neural Processing Letters}, 7:\penalty0 211--219, 1998.
\newblock \doi{10.1023/A:1009670302979}.

\bibitem[{Duygu Arbatli} and {Levent Akin}(1997)]{arbatli1997}
A.~{Duygu Arbatli} and H.~{Levent Akin}.
\newblock Rule extraction from trained neural networks using genetic algorithms.
\newblock \emph{Nonlinear Analysis: Theory, Methods \& Applications}, 30\penalty0 (3):\penalty0 1639--1648, 1997.
\newblock ISSN 0362-546X.
\newblock \doi{https://doi.org/10.1016/S0362-546X(96)00267-2}.
\newblock URL \url{https://www.sciencedirect.com/science/article/pii/S0362546X96002672}.

\bibitem[Frosst and Hinton(2017)]{frosst2017}
N.~Frosst and G.~Hinton.
\newblock Distilling a neural network into a soft decision tree.
\newblock In \emph{CEX workshop at AI*IA 2017 conference}, 2017.
\newblock URL \url{https://arxiv.org/pdf/1711.09784.pdf}.

\bibitem[Fu(1991)]{fu1991}
L.~Fu.
\newblock Rule learning by searching on adapted nets.
\newblock In \emph{9th National Conference on Artificial Intelligence, July 14-19, 1991, Anaheim, California}, pages 590--595, Menlo Park, California, 1991. AAAI Press.

\bibitem[Fu(1994)]{fu1994}
L.~Fu.
\newblock Rule generation from neural networks.
\newblock \emph{IEEE Transactions on Systems, Man, and Cybernetics}, 24\penalty0 (8):\penalty0 1114--1124, 1994.
\newblock \doi{10.1109/21.299696}.

\bibitem[F{\"u}rnkranz et~al.(2012)F{\"u}rnkranz, Gamberger, and Lavra{\v{c}}]{furnkranz2012}
J.~F{\"u}rnkranz, D.~Gamberger, and N.~Lavra{\v{c}}.
\newblock \emph{Foundations of rule learning}.
\newblock Springer Science \& Business Media, 2012.

\bibitem[Gallant(1988)]{gallant1988}
S.~I. Gallant.
\newblock Connectionist expert systems.
\newblock \emph{Commun. ACM}, 31\penalty0 (2):\penalty0 152–169, feb 1988.
\newblock ISSN 0001-0782.
\newblock \doi{10.1145/42372.42377}.
\newblock URL \url{https://doi.org/10.1145/42372.42377}.

\bibitem[Garcez et~al.(2019)Garcez, Gori, Lamb, Serafini, Spranger, and Tran]{garcez2019}
A.~d. Garcez, M.~Gori, L.~C. Lamb, L.~Serafini, M.~Spranger, and S.~N. Tran.
\newblock Neural-symbolic computing: An effective methodology for principled integration of machine learning and reasoning.
\newblock \emph{arXiv preprint arXiv:1905.06088}, 2019.

\bibitem[Goh(1993)]{goh1993}
T.-H. Goh.
\newblock Semantic extraction using neural network modelling and sensitivity analysis.
\newblock In \emph{Proceedings of 1993 International Conference on Neural Networks (IJCNN-93-Nagoya, Japan)}, volume~1, pages 1031--1034, 1993.
\newblock \doi{10.1109/IJCNN.1993.714088}.

\bibitem[Goodfellow et~al.(2016)Goodfellow, Bengio, and Courville]{goodfellow2016}
I.~Goodfellow, Y.~Bengio, and A.~Courville.
\newblock \emph{Deep Learning}.
\newblock MIT Press, 2016.
\newblock \url{http://www.deeplearningbook.org}.

\bibitem[Guidotti et~al.(2018)Guidotti, Monreale, Ruggieri, Pedreschi, Turini, and Giannotti]{guidotti2018}
R.~Guidotti, A.~Monreale, S.~Ruggieri, D.~Pedreschi, F.~Turini, and F.~Giannotti.
\newblock Local rule-based explanations of black box decision systems.
\newblock \emph{arXiv preprint arXiv:1805.10820}, 2018.
\newblock \doi{10.48550/ARXIV.1805.10820}.

\bibitem[Gupta et~al.(1999)Gupta, Park, and Lam]{gupta1999}
A.~Gupta, S.~Park, and S.~M. Lam.
\newblock Generalized analytic rule extraction for feedforward neural networks.
\newblock \emph{IEEE Transactions on Knowledge and Data Engineering}, 11\penalty0 (6):\penalty0 985--991, 1999.
\newblock \doi{10.1109/69.824621}.

\bibitem[Hailesilassie(2016)]{hailesilassie2016}
T.~Hailesilassie.
\newblock Rule extraction algorithm for deep neural networks: A review.
\newblock \emph{arXiv preprint arXiv:1610.05267}, 2016.

\bibitem[Hayward et~al.(1996)Hayward, Tickle, and Diederich]{hayward1996}
R.~Hayward, A.~Tickle, and J.~Diederich.
\newblock Extracting rules for grammar recognition from cascade-2 networks.
\newblock In S.~Wermter, E.~Riloff, and G.~Scheler, editors, \emph{Connectionist, Statistical and Symbolic Approaches to Learning for Natural Language Processing}, pages 48--60, Berlin, Heidelberg, 1996. Springer Berlin Heidelberg.
\newblock ISBN 978-3-540-49738-7.

\bibitem[He et~al.(2020)He, Ma, and Wang]{he2020}
C.~He, M.~Ma, and P.~Wang.
\newblock Extract interpretability-accuracy balanced rules from artificial neural networks: A review.
\newblock \emph{Neurocomputing}, 387:\penalty0 346--358, 2020.

\bibitem[Hruschka(1998)]{hruschka1998}
E.~R. Hruschka.
\newblock Rule extraction from neural networks in data mining applications.
\newblock \emph{WIT Transactions on Information and Communication Technologies}, 22, 1998.
\newblock URL \url{https://www.witpress.com/elibrary/wit-transactions-on-information-and-communication-technologies/22/6942}.

\bibitem[Hruschka and Ebecken(2006)]{Hruschka2006}
E.~R. Hruschka and N.~F. Ebecken.
\newblock Extracting rules from multilayer perceptrons in classification problems: A clustering-based approach.
\newblock \emph{Neurocomputing}, 70\penalty0 (1):\penalty0 384--397, 2006.
\newblock ISSN 0925-2312.
\newblock \doi{10.1016/j.neucom.2005.12.127}.
\newblock URL \url{https://www.sciencedirect.com/science/article/pii/S0925231206000403}.

\bibitem[Huynh and Reggia(2009)]{huynh2009}
T.~Q. Huynh and J.~A. Reggia.
\newblock Improving rule extraction from neural networks by modifying hidden layer representations.
\newblock In \emph{2009 International Joint Conference on Neural Networks}, pages 1316--1321, 2009.
\newblock \doi{10.1109/IJCNN.2009.5178685}.

\bibitem[Huysmans et~al.(2006)Huysmans, Baesens, and Vanthienen]{huysmans2006}
J.~Huysmans, B.~Baesens, and J.~Vanthienen.
\newblock Using rule extraction to improve the comprehensibility of predictive models.
\newblock \emph{K.U. Leuven KBI Working Paper No. 0612}, 2006.
\newblock \doi{http://dx.doi.org/10.2139/ssrn.961358}.

\bibitem[Johansson et~al.(2004)Johansson, K{\"o}nig, and Niklasson]{johansson2004}
U.~Johansson, R.~K{\"o}nig, and L.~Niklasson.
\newblock The truth is in there-rule extraction from opaque models using genetic programming.
\newblock In \emph{Proceedings of the FLAIRS Conference}, pages 658--663, 2004.
\newblock URL \url{https://www.aaai.org/Library/FLAIRS/2004/flairs04-113.php}.

\bibitem[Kahramanli and Allahverdi(2009)]{kahramanli2009}
H.~Kahramanli and N.~Allahverdi.
\newblock Rule extraction from trained adaptive neural networks using artificial immune systems.
\newblock \emph{Expert Systems with Applications}, 36\penalty0 (2, Part 1):\penalty0 1513--1522, 2009.
\newblock ISSN 0957-4174.
\newblock \doi{10.1016/j.eswa.2007.11.024}.
\newblock URL \url{https://www.sciencedirect.com/science/article/pii/S0957417407005878}.

\bibitem[Kaikhah and Doddameti(2006)]{kaikhah2006}
K.~Kaikhah and S.~Doddameti.
\newblock Discovering trends in large datasets using neural networks.
\newblock \emph{Applied Intelligence}, 24:\penalty0 51--60, 2006.
\newblock \doi{10.1007/s10489-006-6929-9}.

\bibitem[Kim(2000)]{kim2000}
H.~Kim.
\newblock Computationally efficient heuristics for if-then rule extraction from feed-forward neural networks.
\newblock In S.~Arikawa and S.~Morishita, editors, \emph{Discovery Science}, pages 170--182, Berlin, Heidelberg, 2000. Springer Berlin Heidelberg.
\newblock ISBN 978-3-540-44418-3.
\newblock \doi{10.1007/3-540-44418-1_14}.

\bibitem[Kitchenham(2004)]{kitchenham2004}
B.~Kitchenham.
\newblock Procedures for performing systematic reviews.
\newblock \emph{Keele, UK, Keele University}, 33\penalty0 (2004):\penalty0 1--26, 2004.

\bibitem[Kohonen(1990)]{kohonen1990}
T.~Kohonen.
\newblock The self-organizing map.
\newblock \emph{Proceedings of the IEEE}, 78\penalty0 (9):\penalty0 1464--1480, 1990.

\bibitem[Krishnan et~al.(1999)Krishnan, Sivakumar, and Bhattacharya]{krishnan1999}
R.~Krishnan, G.~Sivakumar, and P.~Bhattacharya.
\newblock A search technique for rule extraction from trained neural networks.
\newblock \emph{Pattern Recognition Letters}, 20\penalty0 (3):\penalty0 273--280, 1999.
\newblock ISSN 0167-8655.
\newblock \doi{https://doi.org/10.1016/S0167-8655(98)00145-7}.
\newblock URL \url{https://www.sciencedirect.com/science/article/pii/S0167865598001457}.

\bibitem[Lee and Kim(2021)]{lee2021}
H.~Lee and H.~Kim.
\newblock Uncertainty of rules extracted from artificial neural networks.
\newblock \emph{Applied Artificial Intelligence}, 35\penalty0 (8):\penalty0 589--604, 2021.
\newblock \doi{10.1080/08839514.2021.1922845}.
\newblock URL \url{https://doi.org/10.1080/08839514.2021.1922845}.

\bibitem[Liu et~al.(2018)Liu, Wang, and Matwin]{liu2018}
X.~Liu, X.~Wang, and S.~Matwin.
\newblock Improving the interpretability of deep neural networks with knowledge distillation.
\newblock In \emph{2018 IEEE International Conference on Data Mining Workshops (ICDMW)}, pages 905--912, 2018.
\newblock \doi{10.1109/ICDMW.2018.00132}.

\bibitem[L{\"o}fstr{\"o}m et~al.(2004)L{\"o}fstr{\"o}m, Johansson, and Niklasson]{lofstrom2004}
T.~L{\"o}fstr{\"o}m, U.~Johansson, and L.~Niklasson.
\newblock Rule extraction by seeing through the model.
\newblock In N.~R. Pal, N.~Kasabov, R.~K. Mudi, S.~Pal, and S.~K. Parui, editors, \emph{Neural Information Processing}, pages 555--560, Berlin, Heidelberg, 2004. Springer Berlin Heidelberg.
\newblock ISBN 978-3-540-30499-9.
\newblock \doi{10.1007/978-3-540-30499-9_85}.

\bibitem[Lu et~al.(1995)Lu, Setiono, and Liu]{lu1995}
H.~Lu, R.~Setiono, and H.~Liu.
\newblock Neurorule: {A} connectionist approach to data mining.
\newblock In \emph{Proceedings of the 21st VLDB Conference Zurich, Swizerland, 1995}, 1995.
\newblock URL \url{https://arxiv.org/abs/1701.01358}.

\bibitem[Lu et~al.(1996)Lu, Setiono, and Liu]{lu1996}
H.~Lu, R.~Setiono, and H.~Liu.
\newblock Effective data mining using neural networks.
\newblock \emph{IEEE Transactions on Knowledge and Data Engineering}, 8\penalty0 (6):\penalty0 957--961, 1996.
\newblock \doi{10.1109/69.553163}.

\bibitem[Lu et~al.(2006)Lu, Tokinaga, and Ikeda]{lu2006}
J.~Lu, S.~Tokinaga, and Y.~Ikeda.
\newblock Explanatory rule extraction based on the trained neural network and the genetic programming.
\newblock \emph{Journal of the Operations Research Society of Japan}, 49\penalty0 (1):\penalty0 66--82, 2006.
\newblock \doi{10.15807/jorsj.49.66}.

\bibitem[McMillan et~al.(1991{\natexlab{a}})McMillan, Mozer, and Smolensky]{mcmillan1991a}
C.~McMillan, M.~C. Mozer, and P.~Smolensky.
\newblock The connectionist scientist game: rule extraction and refinement in a neural network.
\newblock In \emph{Proceedings of the 13th Annual Conference of the Cognitive Science Society}, pages 424--430, 1991{\natexlab{a}}.

\bibitem[McMillan et~al.(1991{\natexlab{b}})McMillan, Mozer, and Smolensky]{mcmillan1991b}
C.~McMillan, M.~C. Mozer, and P.~Smolensky.
\newblock Rule induction through integrated symbolic and subsymbolic processing.
\newblock In J.~Moody, S.~Hanson, and R.~Lippmann, editors, \emph{Advances in Neural Information Processing Systems}, volume~4, pages 969--976. Morgan-Kaufmann, 1991{\natexlab{b}}.
\newblock URL \url{https://proceedings.neurips.cc/paper/1991/file/cf67355a3333e6e143439161adc2d82e-Paper.pdf}.

\bibitem[Mereani and Howe(2019)]{mereani2019}
F.~A. Mereani and J.~M. Howe.
\newblock Exact and approximate rule extraction from neural networks with boolean features.
\newblock In \emph{Proceedings of the 11th International Joint Conference on Computational Intelligence}, pages 424--433. SCITEPRESS, 2019.
\newblock \doi{10.5220/0008362904240433}.

\bibitem[Milar{\'e} et~al.(2001)Milar{\'e}, de~Carvalho, and Monard]{milare2001}
C.~R. Milar{\'e}, A.~C. de~Carvalho, and M.~C. Monard.
\newblock Extracting rules from neural networks using symbolic algorithms: preliminary results.
\newblock In \emph{Proceedings Fourth International Conference on Computational Intelligence and Multimedia Applications. ICCIMA 2001}, pages 384--388, 2001.
\newblock \doi{10.1109/ICCIMA.2001.970500}.

\bibitem[Odajima et~al.(2006)Odajima, Hayashi, and Setiono]{odajima2006}
K.~Odajima, Y.~Hayashi, and R.~Setiono.
\newblock Greedy rule generation from discrete data and its use in neural network rule extraction.
\newblock In \emph{The 2006 IEEE International Joint Conference on Neural Network Proceedings}, pages 1833--1839, 2006.
\newblock \doi{10.1109/IJCNN.2006.246902}.

\bibitem[{\"O}zbak{\i}r et~al.(2008){\"O}zbak{\i}r, Baykaso{\u{g}}lu, and Kulluk]{ozbakir2008}
L.~{\"O}zbak{\i}r, A.~Baykaso{\u{g}}lu, and S.~Kulluk.
\newblock Rule extraction from neural networks via ant colony algorithm for data mining applications.
\newblock In V.~Maniezzo, R.~Battiti, and J.-P. Watson, editors, \emph{Learning and Intelligent Optimization}, pages 177--191, Berlin, Heidelberg, 2008. Springer Berlin Heidelberg.
\newblock ISBN 978-3-540-92695-5.
\newblock \doi{10.1007/978-3-540-92695-5_14}.

\bibitem[{\"O}zbak{\i}r et~al.(2009){\"O}zbak{\i}r, Baykaso{\u{g}}lu, Kulluk, and Yapıcı]{ozbakir2009}
L.~{\"O}zbak{\i}r, A.~Baykaso{\u{g}}lu, S.~Kulluk, and H.~Yapıcı.
\newblock Taco-miner: An ant colony based algorithm for rule extraction from trained neural networks.
\newblock \emph{Expert Systems with Applications}, 36\penalty0 (10):\penalty0 12295--12305, 2009.
\newblock ISSN 0957-4174.
\newblock \doi{10.1016/j.eswa.2009.04.058}.
\newblock URL \url{https://www.sciencedirect.com/science/article/pii/S0957417409004084}.

\bibitem[{\"O}zbak{\i}r et~al.(2010){\"O}zbak{\i}r, Baykaso{\u{g}}lu, and Kulluk]{ozbakir2010}
L.~{\"O}zbak{\i}r, A.~Baykaso{\u{g}}lu, and S.~Kulluk.
\newblock A soft computing-based approach for integrated training and rule extraction from artificial neural networks: Difaconn-miner.
\newblock \emph{Applied Soft Computing}, 10\penalty0 (1):\penalty0 304--317, 2010.
\newblock ISSN 1568-4946.
\newblock \doi{10.1016/j.asoc.2009.08.008}.
\newblock URL \url{https://www.sciencedirect.com/science/article/pii/S1568494609001318}.

\bibitem[Palade et~al.(2001)Palade, Neagu, and Patton]{palade2001}
V.~Palade, D.-C. Neagu, and R.~J. Patton.
\newblock Interpretation of trained neural networks by rule extraction.
\newblock In B.~Reusch, editor, \emph{Computational Intelligence. Theory and Applications}, pages 152--161, Berlin, Heidelberg, 2001. Springer Berlin Heidelberg.
\newblock ISBN 978-3-540-45493-9.
\newblock \doi{10.1007/3-540-45493-4_20}.

\bibitem[Pearl(2009)]{pearl2009}
J.~Pearl.
\newblock \emph{Causality}.
\newblock Cambridge university press, 2009.

\bibitem[Rasamoelina et~al.(2020)Rasamoelina, Adjailia, and Sin{\v{c}}{\'a}k]{rasamoelina2020}
A.~D. Rasamoelina, F.~Adjailia, and P.~Sin{\v{c}}{\'a}k.
\newblock A review of activation function for artificial neural network.
\newblock In \emph{2020 IEEE 18th World Symposium on Applied Machine Intelligence and Informatics (SAMI)}, pages 281--286. IEEE, 2020.

\bibitem[Raschka(2018)]{raschka2018}
S.~Raschka.
\newblock Model evaluation, model selection, and algorithm selection in machine learning.
\newblock \emph{arXiv preprint arXiv:1811.12808}, 2018.

\bibitem[Ribeiro et~al.(2016)Ribeiro, Singh, and Guestrin]{ribeiro2016}
M.~T. Ribeiro, S.~Singh, and C.~Guestrin.
\newblock " why should i trust you?" explaining the predictions of any classifier.
\newblock In \emph{Proceedings of the 22nd ACM SIGKDD international conference on knowledge discovery and data mining}, pages 1135--1144, 2016.

\bibitem[Ribeiro et~al.(2018)Ribeiro, Singh, and Guestrin]{ribeiro2018}
M.~T. Ribeiro, S.~Singh, and C.~Guestrin.
\newblock Anchors: High-precision model-agnostic explanations.
\newblock \emph{Proceedings of the AAAI Conference on Artificial Intelligence}, 32\penalty0 (1), Apr. 2018.
\newblock \doi{10.1609/aaai.v32i1.11491}.
\newblock URL \url{https://ojs.aaai.org/index.php/AAAI/article/view/11491}.

\bibitem[Saito and Nakano(1988)]{saito1988}
K.~Saito and R.~Nakano.
\newblock Medical diagnostic expert system based on pdp model.
\newblock In \emph{IEEE 1988 International Conference on Neural Networks}, volume~1, pages 255--262, 1988.
\newblock \doi{10.1109/ICNN.1988.23855}.

\bibitem[Saito and Nakano(1990)]{saito1990}
K.~Saito and R.~Nakano.
\newblock Automatic extraction of classification rules.
\newblock In \emph{International Neural Network Conference, July 9-13, 1990, Palais des Congres, Paris, France}, pages 379--382, 1990.

\bibitem[Saito and Nakano(2002)]{saito2002}
K.~Saito and R.~Nakano.
\newblock Extracting regression rules from neural networks.
\newblock \emph{Neural Networks}, 15\penalty0 (10):\penalty0 1279--1288, 2002.
\newblock ISSN 0893-6080.
\newblock \doi{10.1016/S0893-6080(02)00089-8}.
\newblock URL \url{https://www.sciencedirect.com/science/article/pii/S0893608002000898}.

\bibitem[Sato and Tsukimoto(2001)]{sato2001}
M.~Sato and H.~Tsukimoto.
\newblock Rule extraction from neural networks via decision tree induction.
\newblock In \emph{IJCNN'01. International Joint Conference on Neural Networks. Proceedings (Cat. No.01CH37222)}, volume~3, pages 1870--1875, 2001.
\newblock \doi{10.1109/IJCNN.2001.938448}.

\bibitem[Schmitz et~al.(1999)Schmitz, Aldrich, and Gouws]{schmitz1999}
G.~P. Schmitz, C.~Aldrich, and F.~S. Gouws.
\newblock Ann-dt: an algorithm for extraction of decision trees from artificial neural networks.
\newblock \emph{IEEE Transactions on Neural Networks}, 10\penalty0 (6):\penalty0 1392--1401, 1999.
\newblock \doi{10.1109/72.809084}.

\bibitem[Sestito and Dillon(1991)]{sestito1991}
S.~Sestito and T.~Dillon.
\newblock Using single-layered neural networks for the extraction of conjunctive rules and hierarchical classifications.
\newblock \emph{Applied Intelligence}, 1:\penalty0 157--173, 1991.
\newblock \doi{10.1007/BF00058881}.

\bibitem[Sestito and Dillon(1993)]{sestito1993}
S.~Sestito and T.~Dillon.
\newblock Knowledge acquisition of conjunctive rules using multilayered neural networks.
\newblock \emph{International Journal of Intelligent Systems}, 8\penalty0 (7):\penalty0 779--805, 1993.
\newblock \doi{https://doi.org/10.1002/int.4550080704}.
\newblock URL \url{https://onlinelibrary.wiley.com/doi/abs/10.1002/int.4550080704}.

\bibitem[Sestito and Dillon(1990)]{sestito1990}
S.~Sestito and T.~S. Dillon.
\newblock Machine learning using single-layered and multi-layered neural networks.
\newblock In \emph{Proceedings of the 2nd International IEEE Conference on Tools for Artificial Intelligence}, pages 269--275, 1990.
\newblock \doi{10.1109/TAI.1990.130346}.

\bibitem[Sethi and Yoo(1994)]{sethi1994}
I.~K. Sethi and J.~H. Yoo.
\newblock Symbolic approximation of feedforward neural networks.
\newblock In E.~S. GELSEMA and L.~S. KANAL, editors, \emph{Pattern Recognition in Practice IV}, volume~16 of \emph{Machine Intelligence and Pattern Recognition}, pages 313--324. North-Holland, 1994.
\newblock \doi{https://doi.org/10.1016/B978-0-444-81892-8.50032-8}.
\newblock URL \url{https://www.sciencedirect.com/science/article/pii/B9780444818928500328}.

\bibitem[Sethi et~al.(2012)Sethi, Mishra, and Mishra]{sethi2012}
K.~K. Sethi, D.~K. Mishra, and B.~Mishra.
\newblock Kdruleex: A novel approach for enhancing user comprehensibility using rule extraction.
\newblock In \emph{2012 Third International Conference on Intelligent Systems Modelling and Simulation}, pages 55--60, 2012.
\newblock \doi{10.1109/ISMS.2012.116}.

\bibitem[Setiono(1997)]{setiono1997b}
R.~Setiono.
\newblock {Extracting Rules from Neural Networks by Pruning and Hidden-Unit Splitting}.
\newblock \emph{Neural Computation}, 9\penalty0 (1):\penalty0 205--225, 01 1997.
\newblock ISSN 0899-7667.
\newblock \doi{10.1162/neco.1997.9.1.205}.
\newblock URL \url{https://doi.org/10.1162/neco.1997.9.1.205}.

\bibitem[Setiono and Leow(2000)]{setiono2000}
R.~Setiono and W.~K. Leow.
\newblock Fernn: An algorithm for fast extraction of rules from neural networks.
\newblock \emph{Applied Intelligence}, 12:\penalty0 15--25, 2000.
\newblock \doi{10.1023/A:1008307919726}.

\bibitem[Setiono and Liu(1996)]{setiono1996}
R.~Setiono and H.~Liu.
\newblock Symbolic representation of neural networks.
\newblock \emph{Computer}, 29\penalty0 (3):\penalty0 71--77, 1996.
\newblock \doi{10.1109/2.485895}.

\bibitem[Setiono and Liu(1997)]{setiono1997a}
R.~Setiono and H.~Liu.
\newblock Neurolinear: From neural networks to oblique decision rules.
\newblock \emph{Neurocomputing}, 17\penalty0 (1):\penalty0 1--24, 1997.
\newblock ISSN 0925-2312.
\newblock \doi{https://doi.org/10.1016/S0925-2312(97)00038-6}.
\newblock URL \url{https://www.sciencedirect.com/science/article/pii/S0925231297000386}.

\bibitem[Setiono and Liu(August 20--25,1995)]{setiono1995}
R.~Setiono and H.~Liu.
\newblock Understanding neural networks via rule extraction.
\newblock In \emph{Proceedings of the Fourteenth International Joint Conference on Artificial Intelligence (IJCAI), Montreal, Quebec, Canada}, pages 480--485, August 20--25,1995.
\newblock URL \url{https://www.ijcai.org/Proceedings/95-1/Papers/063.pdf}.

\bibitem[Setiono and Thong(2004)]{setiono2004}
R.~Setiono and J.~Y. Thong.
\newblock An approach to generate rules from neural networks for regression problems.
\newblock \emph{European Journal of Operational Research}, 155\penalty0 (1):\penalty0 239--250, 2004.
\newblock URL \url{https://ssrn.com/abstract=3766019}.

\bibitem[Setiono et~al.(2002)Setiono, Leow, and Zurada]{setiono2002}
R.~Setiono, W.~K. Leow, and J.~M. Zurada.
\newblock Extraction of rules from artificial neural networks for nonlinear regression.
\newblock \emph{IEEE Transactions on Neural Networks}, 13\penalty0 (3):\penalty0 564--577, 2002.
\newblock \doi{10.1109/TNN.2002.1000125}.

\bibitem[Setiono et~al.(2008)Setiono, Baesens, and Mues]{setiono2008}
R.~Setiono, B.~Baesens, and C.~Mues.
\newblock Recursive neural network rule extraction for data with mixed attributes.
\newblock \emph{IEEE Transactions on Neural Networks}, 19\penalty0 (2):\penalty0 299--307, 2008.
\newblock \doi{10.1109/TNN.2007.908641}.

\bibitem[Shams et~al.(2021)Shams, Dimanov, Kola, Simidjievski, Terre, Scherer, Matja{\v s}ec, Abraham, Jamnik, and Li{\`o}]{shams2021}
Z.~Shams, B.~Dimanov, S.~Kola, N.~Simidjievski, H.~A. Terre, P.~Scherer, U.~Matja{\v s}ec, J.~Abraham, M.~Jamnik, and P.~Li{\`o}.
\newblock Rem: An integrative rule extraction methodology for explainable data analysis in healthcare.
\newblock \emph{medRxiv}, 2021.
\newblock \doi{10.1101/2021.01.25.21250459}.
\newblock URL \url{https://www.medrxiv.org/content/early/2021/04/28/2021.01.25.21250459}.

\bibitem[Singh et~al.(2016)Singh, Ribeiro, and Guestrin]{singh2016}
S.~Singh, M.~T. Ribeiro, and C.~Guestrin.
\newblock Programs as black-box explanations, 2016.
\newblock URL \url{https://arxiv.org/abs/1611.07579}.

\bibitem[Taha and Ghosh(1999)]{taha1999}
I.~Taha and J.~Ghosh.
\newblock Symbolic interpretation of artificial neural networks.
\newblock \emph{IEEE Transactions on Knowledge and Data Engineering}, 11\penalty0 (3):\penalty0 448--463, 1999.
\newblock \doi{10.1109/69.774103}.

\bibitem[Thiagarajan et~al.(2016)Thiagarajan, Kailkhura, Sattigeri, and Ramamurthy]{thiagarajan2016}
J.~J. Thiagarajan, B.~Kailkhura, P.~Sattigeri, and K.~N. Ramamurthy.
\newblock Treeview: Peeking into deep neural networks via feature-space partitioning.
\newblock \emph{arXiv preprint arXiv:1611.07429}, 2016.
\newblock URL \url{https://arxiv.org/abs/1611.07429}.

\bibitem[Thrun(1993)]{thrun1993}
S.~Thrun.
\newblock Extracting provably correct rules from artificial neural networks.
\newblock Technical report, University of Bonn, Institut fur Informatik III, 1993.
\newblock URL \url{https://citeseerx.ist.psu.edu/viewdoc/download?doi=10.1.1.2.2110&rep=rep1&type=pdf}.

\bibitem[Thrun(1994)]{thrun1994}
S.~Thrun.
\newblock Extracting rules from artificial neural networks with distributed representations.
\newblock In G.~Tesauro, D.~Touretzky, and T.~Leen, editors, \emph{Advances in Neural Information Processing Systems}, volume~7. MIT Press, 1994.
\newblock URL \url{https://proceedings.neurips.cc/paper/1994/file/bea5955b308361a1b07bc55042e25e54-Paper.pdf}.

\bibitem[Tickle et~al.(1994)Tickle, Orlowski, and Diederich]{tickle1994}
A.~Tickle, M.~Orlowski, and J.~Diederich.
\newblock Dedec: Decision detection by rule extraction from neural networks. queensland university of technology.
\newblock \emph{Neurocomputing Research Center QUT NRC}, 1994.

\bibitem[Tickle et~al.(1998)Tickle, Andrews, Golea, and Diederich]{tickle1998}
A.~B. Tickle, R.~Andrews, M.~Golea, and J.~Diederich.
\newblock The truth will come to light: Directions and challenges in extracting the knowledge embedded within trained artificial neural networks.
\newblock \emph{IEEE Transactions on Neural Networks}, 9\penalty0 (6):\penalty0 1057--1068, 1998.

\bibitem[Towell and Shavlik(1993)]{towell1993}
G.~G. Towell and J.~W. Shavlik.
\newblock Extracting refined rules from knowledge-based neural networks.
\newblock \emph{Machine Learning}, 13:\penalty0 71--101, 1993.
\newblock \doi{10.1007/BF00993103}.

\bibitem[Townsend et~al.(2019)Townsend, Chaton, and Monteiro]{townsend2019}
J.~Townsend, T.~Chaton, and J.~M. Monteiro.
\newblock Extracting relational explanations from deep neural networks: A survey from a neural-symbolic perspective.
\newblock \emph{IEEE transactions on neural networks and learning systems}, 31\penalty0 (9):\penalty0 3456--3470, 2019.

\bibitem[Tsukimoto(2000)]{tsukimoto2000}
H.~Tsukimoto.
\newblock Extracting rules from trained neural networks.
\newblock \emph{IEEE Transactions on Neural Networks}, 11\penalty0 (2):\penalty0 377--389, 2000.
\newblock \doi{10.1109/72.839008}.

\bibitem[UZUN et~al.(2016)UZUN, ARIKAN, and TEZEL]{uzun2016}
Y.~UZUN, H.~ARIKAN, and G.~TEZEL.
\newblock Rule extraction from training artificial neural network using variable neighbourhood search for wisconsin breast cancer.
\newblock \emph{Journal of Multidisciplinary Engineering Science and Technology}, 3\penalty0 (8), 2016.
\newblock URL \url{http://www.jmest.org/wp-content/uploads/JMESTN42351743.pdf}.

\bibitem[Vilone and Longo(2020)]{vilone2020}
G.~Vilone and L.~Longo.
\newblock Explainable artificial intelligence: a systematic review, 2020.

\bibitem[Wu et~al.(2018)Wu, Hughes, Parbhoo, Zazzi, Roth, and Doshi-Velez]{wu2018}
M.~Wu, M.~Hughes, S.~Parbhoo, M.~Zazzi, V.~Roth, and F.~Doshi-Velez.
\newblock Beyond sparsity: Tree regularization of deep models for interpretability.
\newblock In \emph{Proceedings of the AAAI conference on artificial intelligence}, volume~32, 2018.
\newblock \doi{10.1609/aaai.v32i1.11501}.
\newblock URL \url{https://ojs.aaai.org/index.php/AAAI/article/view/11501}.

\bibitem[Yedjour(2020)]{yedjour2020}
D.~Yedjour.
\newblock Extracting classification rules from artificial neural network trained with discretized inputs.
\newblock \emph{Neural Processing Letters}, 52:\penalty0 2469--2491, 2020.
\newblock \doi{10.1007/s11063-020-10357-x}.

\bibitem[Yedjour and Benyettou(2018)]{yedjour2018}
D.~Yedjour and A.~Benyettou.
\newblock Symbolic interpretation of artificial neural networks based on multiobjective genetic algorithms and association rules mining.
\newblock \emph{Applied Soft Computing}, 72:\penalty0 177--188, 2018.
\newblock ISSN 1568-4946.
\newblock \doi{10.1016/j.asoc.2018.08.007}.
\newblock URL \url{https://www.sciencedirect.com/science/article/pii/S1568494618304551}.

\bibitem[Zarlenga et~al.(2021)Zarlenga, Shams, and Jamnik]{zarlenga2021}
M.~E. Zarlenga, Z.~Shams, and M.~Jamnik.
\newblock Efficient decompositional rule extraction for deep neural networks, 2021.
\newblock URL \url{https://arxiv.org/abs/2111.12628}.

\bibitem[Zhou et~al.(2000)Zhou, Chen, and Chen]{zhou2000}
Z.-H. Zhou, S.-F. Chen, and Z.-Q. Chen.
\newblock A statistics based approach for extracting priority rules from trained neural networks.
\newblock In \emph{Proceedings of the IEEE-INNS-ENNS International Joint Conference on Neural Networks. IJCNN 2000. Neural Computing: New Challenges and Perspectives for the New Millennium}, volume~3, pages 401--406, 2000.
\newblock \doi{10.1109/IJCNN.2000.861337}.

\bibitem[Zilke et~al.(2016)Zilke, Loza~Menc{\'i}a, and Janssen]{zilke2016}
J.~R. Zilke, E.~Loza~Menc{\'i}a, and F.~Janssen.
\newblock Deepred -- rule extraction from deep neural networks.
\newblock In T.~Calders, M.~Ceci, and D.~Malerba, editors, \emph{Discovery Science}, pages 457--473, Cham, 2016. Springer International Publishing.
\newblock ISBN 978-3-319-46307-0.
\newblock \doi{10.1007/978-3-319-46307-0_29}.

\end{thebibliography}





\end{document}